%% file: PaperForReview_jcy3.tex
\documentclass[10pt,twocolumn,letterpaper]{article}

\usepackage[pagenumbers]{cvpr}      

\usepackage{graphicx}
\usepackage{amsmath}
\usepackage{amssymb}
\usepackage{booktabs}
\usepackage[utf8]{inputenc} 
\usepackage[T1]{fontenc}    
\usepackage{url}            
\usepackage{booktabs}       
\usepackage{amsfonts}       
\usepackage{nicefrac}       
\usepackage{microtype}      
\usepackage{xcolor}         
\usepackage[linesnumbered,ruled,vlined]{algorithm2e}
\usepackage{array}
\usepackage{amssymb}
\usepackage{latexsym}
\usepackage{amsmath,amssymb}
\usepackage{graphicx}
\usepackage{multirow}
\usepackage{adjustbox}
\usepackage{makecell}
\usepackage[nottoc]{tocbibind}\usepackage{url}
\usepackage{wrapfig}

\SetCommentSty{mycommfont}
\newlength{\commentWidth}
\usepackage[accsupp]{axessibility} 

\usepackage[pagebackref,breaklinks,colorlinks]{hyperref}

\newcommand{\textred}[1] {\textcolor{red}{#1}} 
\newcommand\blfootnote[1]{%
  \begingroup
  \renewcommand\thefootnote{}\footnote{#1}%
  \addtocounter{footnote}{-1}%
  \endgroup
}

\input{math_commands}

\input{macros}

\def\vf{{\bm{f}}}

\def\vx{{\bm{x}}}

\renewcommand{\add}[1] {\textcolor{black}{#1}} 

\usepackage[capitalize]{cleveref}
\crefname{section}{Sec.}{Secs.}
\Crefname{section}{Section}{Sections}
\Crefname{table}{Table}{Tables}
\crefname{table}{Tab.}{Tabs.}

\def\cvprPaperID{10095} 
\def\confName{CVPR}
\def\confYear{2022}

\begin{document}

\title{DiffusionCLIP: Text-Guided Diffusion Models for Robust Image Manipulation}

\author{Gwanghyun Kim$^1$ \qquad Taesung Kwon$^1$ \qquad Jong Chul Ye$^{2,1}$ \\
Dept. of Bio and Brain Engineering$^1$,  Kim Jaechul Graduate School of AI$^2$ \\
Korea Advanced Institute of Science and Technology (KAIST), Daejeon, Korea \\
{\tt\small \{gwang.kim, star.kwon, jong.ye\}@kaist.ac.kr}
}

\twocolumn[{%
\renewcommand\twocolumn[1][]{#1}%
\maketitle
\begin{center}
    \centering
    \captionsetup{type=figure}
    \vspace{-2.5em}
    \includegraphics[width=0.94\textwidth]{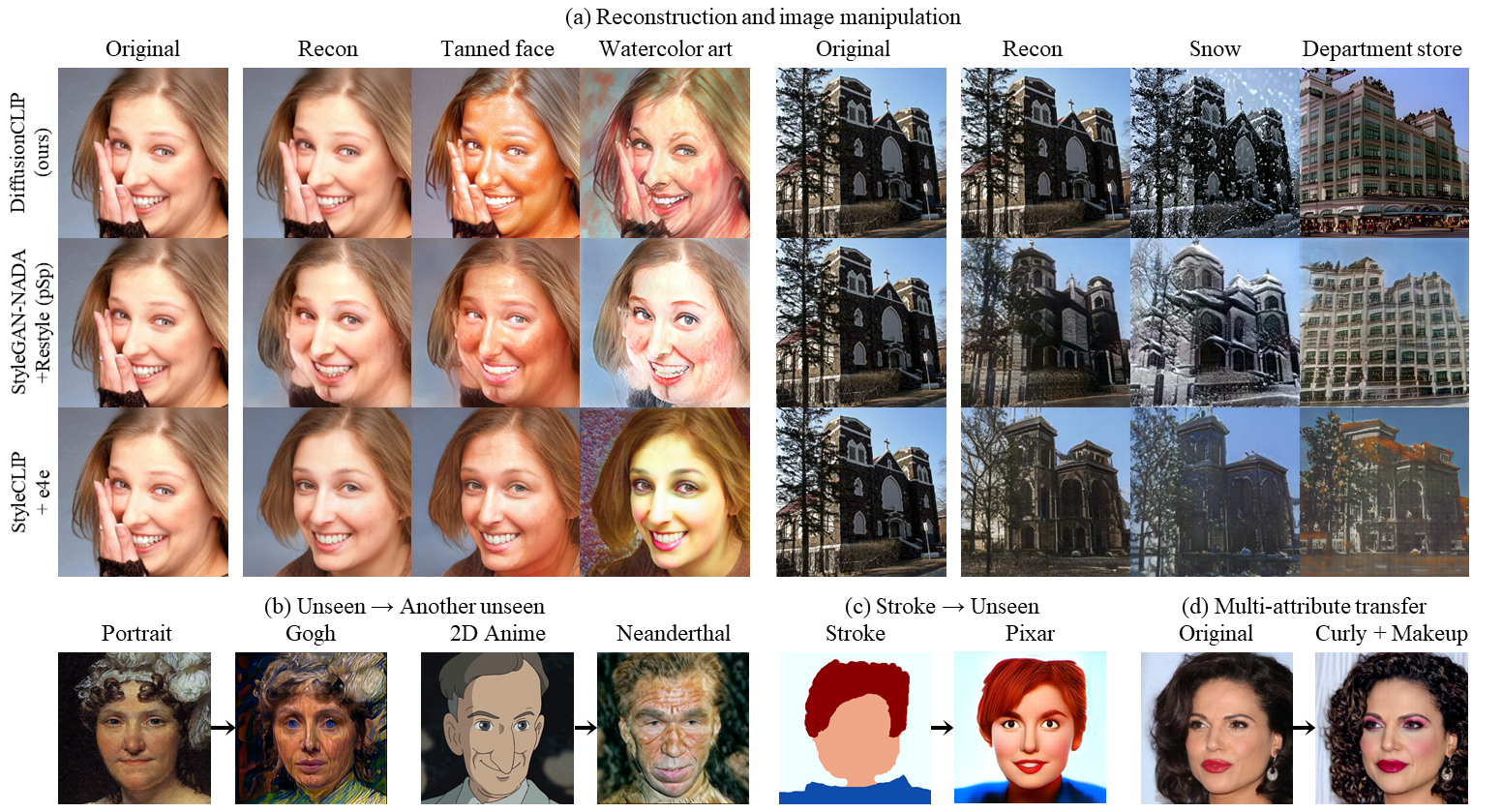}
    \vspace{-0.8em}
    \captionof{figure}{DiffusionCLIP enables faithful text-driven manipulation of real images by (a) preserving important details when the state-of-the-art GAN inversion-based methods fail. Other novel applications include (b) image translation between two unseen domains, (c) stroke-conditioned image synthesis to an unseen domain, and (d) multi-attribute transfer.}
    \vspace{-0.5em}
    \label{fig:1_overview}
\end{center}%
}]

\blfootnote{{This research was supported by Field-oriented Technology Development Project for Customs Administration through the National Research Foundation of Korea(NRF) funded by the Ministry of Science \& ICT and Korea Customs Service (NRF-2021M3I1A1097938), and supported by the Institute of Information \& communications Technology Planning \& Evaluation (IITP) grant funded by the Korea government (MSIT) 
(No.2019-0-00075, Artificial Intelligence Graduate School Program (KAIST)).}}

\vspace{-1em}
\begin{abstract}
Recently, GAN inversion methods combined with Contrastive Language-Image Pretraining (CLIP) enables zero-shot image manipulation guided by text prompts. However, their applications to diverse real images are still difficult due to the limited GAN inversion capability. Specifically, these approaches often have difficulties in reconstructing images with novel poses, views, and highly variable contents compared to the training data, altering object identity, or producing unwanted image artifacts. To mitigate these problems and enable faithful manipulation of real images, we propose a novel method, dubbed DiffusionCLIP, that performs text-driven image manipulation using diffusion models. Based on full inversion capability and high-quality image generation power of recent diffusion models, our method performs zero-shot image manipulation successfully even between unseen domains and takes another step towards general application by manipulating images from a widely varying ImageNet dataset. Furthermore, we propose a novel noise combination method that allows straightforward multi-attribute manipulation. Extensive experiments and human evaluation confirmed robust and superior manipulation performance of our methods compared to the existing baselines. Code is available at \href{https://github.com/gwang-kim/DiffusionCLIP.git}{https://github.com/gwang-kim/DiffusionCLIP.git}
    \end{abstract}


\vspace{-1.5em}

\section{Introduction}
\label{sec:intro}

Recently, GAN inversion methods \cite{brock2016neural, richardson2021encoding, tov2021designing, bau2020semantic, abdal2019image2stylegan, abdal2020image2stylegan++, alaluf2021restyle} combined with Contrastive Language-Image Pretraining (CLIP) \cite{radford2021learning} has become popular
thanks to their ability for zero-shot image manipulation guided by text prompts  \cite{patashnik2021styleclip, gal2021stylegan}. Nevertheless, its real-world application on diverse types of images is still tricky due to the limited GAN inversion performance. 

Specifically, successful manipulation of images should convert the image attribute to that of the target without unintended changes of the input content. Unfortunately,
the current state-of-the-art (SOTA) encoder-based GAN inversion approaches \cite{tov2021designing, richardson2021encoding, alaluf2021restyle} often fail to reconstruct images with novel poses, views, and details. For example, in the left panel of Fig.~\ref{fig:1_overview}(a), e4e \cite{tov2021designing} and ReStyle \cite{alaluf2021restyle} with pSp encoder \cite{richardson2021encoding} fail to reconstruct unexpected hand on the cheek, inducing the unintended change. This is because they have rarely seen
such faces with hands during the training phase.
This issue becomes even worse in the case of images from a dataset with high variance such as church images in LSUN-Church~\cite{yu15lsun} and ImageNet~\cite{ILSVRC15} dataset. As shown in the right panel of Fig.~\ref{fig:1_overview}(a) for the conversion to a department store, existing GAN inversion methods produce artificial architectures that can be perceived as different buildings.

Recently, diffusion models such as denoising diffusion probabilistic models (DDPM) \cite{ho2020denoising, sohl2015deep} and score-based generative models \cite{song2019generative, song2020score} have achieved great successes in image generation  tasks \cite{ho2020denoising,song2020denoising,song2020score,jolicoeur2020adversarial}. The latest works \cite{song2020score, dhariwal2021diffusion} have demonstrated even higher quality of image synthesis performance compared to
variational autoencoders (VAEs) \cite{kingma2013auto, razavi2019generating, oord2017neural}, flows \cite{rezende2015variational, dinh2016density, kingma2018glow}, auto-regressive models \cite{menick2018generating, van2016pixel}, and generative adversarial networks (GANs) \cite{goodfellow2014generative, karras2020analyzing, karras2019style,brock2018large}.
 Furthermore, a recent
 denoising diffusion implicit models (DDIM) \cite{song2020denoising} further accelerates  sampling procedure
and enables nearly perfect inversion \cite{dhariwal2021diffusion}.

Inspired by this, here we propose a novel DiffusionCLIP - a CLIP-guided robust image manipulation method by diffusion models. Here, an input image is first converted to the latent noises through a forward diffusion. In the case of DDIM, the latent noises can be then inverted nearly perfectly to the original image using a reverse diffusion if the score function for the reverse diffusion is retained the same as that of the forward diffusion. Therefore, the key idea of DiffusionCLIP is to fine-tune the score function in the reverse diffusion process using a CLIP loss that controls the attributes of the generated image based on the text prompts.

Accordingly,  DiffusionCLIP can successfully perform image manipulation both in the trained and unseen domain (Fig.~\ref{fig:1_overview}(a)). We can even translate the image from an unseen domain into another unseen domain (Fig.~\ref{fig:1_overview}(b)), or generate images in an unseen domain from the strokes (Fig.~\ref{fig:1_overview}(c)). 
  Moreover, by simply combining the noise predicted from several fine-tuned models, multiple attributes can be changed simultaneously through only one sampling process (Fig.~\ref{fig:1_overview}(d)).
{Furthermore, DiffsuionCLIP takes another step towards general application by manipulating images from a widely varying ImageNet~\cite{ILSVRC15} dataset (Fig.~\ref{fig:6_more_results}), which has been rarely explored with GAN-inversion due to its inferior reconstruction.~\cite{daras2020your, bora2017compressed}}

Additionally, we propose a systematic approach to find the optimal sampling conditions that lead to high quality and speedy image manipulation. Qualitative comparison and human evaluation results demonstrate that our method can provide robust and accurate image manipulation, outperforming SOTA baselines.

\section{Related Works}
\label{sec:background}
\subsection{Diffusion Models}
Diffusion probabilistic models \cite{ho2020denoising, sohl2015deep} are a type of latent variable models that consist of a forward diffusion process and a reverse diffusion process.
The forward process is a Markov chain where noise is gradually added to the data when sequentially sampling the latent variables $\xb_t$ for $t = 1, \cdots , T$. Each step in the forward process is a Gaussian transition $q(\xb_t \mid \x_{t-1}) :=\mathcal{N}(\sqrt{1-\beta_t}\xb_{t-1}, \beta_t\mathbf{I})$, where $\{\beta_t\}^T_{t=0}$ are fixed or learned variance schedule. The resulting latent variable $\xb_t$ can be expressed as:
\begin{align}
\label{eq:forward_ddpm}
    \xb_t = \sqrt{\alpha_t}\xb_0  + \sqrt{1 - \alpha_t}\bm{\wb}, \ \  \ \bm{\wb} \sim \mathcal{N} (\mathbf{0,I}),
\end{align}
where $\alpha_t := \prod_{s=1}^{t} {(1-\beta_s)}.$ The reverse process $q(\xb_{t-1} \mid \x_{t})$ is  parametrized by another Gaussian transition  $p_\theta(\xb_{t-1}\mid\xb_{t}) := \mathcal{N}(\xb_{t-1}; \bm{\mu}_\theta(\xb_t, t), \ \sigma_{\theta}(\xb_t, t)\mathbf{I})$. ${\bm{\mu}_\theta(\xb_t, t)}$ can be decomposed into the linear combination of $\xb_t$ and a noise approximation model $\epsilonb_\theta(\xb_t, t)$, which can be learned by solving the optimization problem as follows:
 \begin{align}
     \min_{\theta}\mathbb{E}_{\xb_0 \sim q(\xb_0), \bm{\wb} \sim \mathcal{N}(\mathbf{0,I}), t} ||\bm{\wb}-\epsilonb_\theta(\xb_t, t) ||^2_2.
\end{align}
After training  $\epsilonb_\theta(\xb, t)$, the data is sampled using following reverse diffusion process:
\begin{align}
    \label{eq:reverse_ddpm}
    \xb_{t-1} = \frac{1}{\sqrt{1-\beta_t}}\left(\xb_t - \frac{\beta_t}{\sqrt{1-\alpha_t}} \epsilonb_\theta(\xb_t, t)\right) + \sigma_t\zb,
\end{align}
where $\bm{\zb} \sim \mathcal{N} (\mathbf{0,I})$. It was found that the sampling process of DDPM  corresponds to that of the score-based generative models \cite{song2019generative, song2020score} with the following relationship:
\begin{align}
    \label{score_ddpm_relation}
    \epsilonb_\theta(\xb_t, t) = -\sqrt{1-\alpha_t}\nabla_{\xb_t}\log p_{\theta}(\xb_t).
\end{align}

Meanwhile, \cite{song2020denoising} proposed an alternative non-Markovian noising process that has the same forward marginals as DDPM 
but has a distinct sampling process as follows:
\small
\begin{gather}
\xb_{t-1} = \sqrt{\alpha_{t-1}}\fb_\theta(\xb_{t}, t) +  \sqrt{1 - \alpha_{t-1} - \sigma^2_t}{\epsilonb}_{\theta}(\xb_{t}, t) + \sigma^2_t\zb,
\label{eq:ddim_original}
\end{gather}
\normalsize 
where, $\bm{\zb} \sim \mathcal{N} (\mathbf{0,I})$ and $\vf_\theta(\xb_{t}, t)$ is a the prediction of $\xb_0$ at $t$ given $\xb_t$ and $\epsilonb_\theta(\vx_t, t)$:
\begin{align}\label{eq:f}
  \vf_\theta(\xb_{t}, t):= \frac{\xb_{t} - \sqrt{1-\alpha_t}\epsilonb_{\theta}(\xb_t,t)}{\sqrt{\alpha_{t}}}.
 \end{align}
This sampling allows using different samplers by changing the variance of the noise $\sigma_t$. Especially, by setting this noise to 0, which is a DDIM sampling process  \cite{song2020denoising}, the sampling process becomes deterministic, enabling full inversion of the latent variables into the original images with significantly fewer steps \cite{song2020denoising, dhariwal2021diffusion}. 
In fact, DDIM can be considered as an Euler method to solve an ordinary differential equation (ODE) by rewriting Eq. \ref{eq:ddim_original} as follows:
\small
\begin{align}
\sqrt{\frac{1}{\alpha_{t-1}}}\xb_{t-1}  - \sqrt{\frac{1}{\alpha_{t}}}\xb_{t}  = \left(\sqrt{\frac{1}{\alpha_{t-1}}-1} - \sqrt{\frac{1}{\alpha_{t}}-1}\right) \epsilonb_\theta(\xb_t, t).
\end{align}
\normalsize
{For mathematical details, see Supplementary Section \textred{A}.}


\subsection{CLIP Guidance for Image Manipulation}
\label{sec:CLIP}

CLIP \cite{radford2021learning} was proposed to efficiently learn visual concepts with natural language supervision. In CLIP, a text encoder and an image encoder are pretrained to identify which texts are matched with which images in the dataset. 
Accordingly,
we use a pretrained CLIP model for our text-driven image manipulation.
 
 To effectively extract knowledge from CLIP, two different losses have been proposed: a global target loss \cite{patashnik2021styleclip}, and local directional loss  \cite{gal2021stylegan}.
The global CLIP loss  tries to minimize the cosine distance in the CLIP space between the generated image and a given target text as follows:
\begin{equation}
    \mathcal{L}_{\text{global}}(\xb_\text{gen}, {y_\text{tar}}) = D_{\text{CLIP}}(\xb_\text{gen}, {y_\text{tar}}),
\end{equation}
where {$y_\text{tar}$} is a text description of a target, $\xb_\text{gen}$ denotes the generated image, and $D_\text{CLIP}$ returns a cosine distance in the CLIP space between
their encoded vectors.
On the other hand, the local directional loss  \cite{gal2021stylegan} is designed to alleviate the issues of global CLIP loss such as low diversity and susceptibility to adversarial attacks. The local directional CLIP loss induces the direction between the embeddings of the reference and generated images to be aligned with the direction between the embeddings of a pair of reference and target texts in the CLIP space as follows:
 \begin{eqnarray}
     \mathcal{L}_{\text{direction}}\left(\xb_{\text{gen}},{y_{\text{tar}}}; \xb_{\text{ref}},{y_{\text{ref}}}\right): = 1 - \frac{\langle \Delta I, \Delta T\rangle}{\|\Delta I\|\|\Delta T\|},
     \end{eqnarray}
 where     \begin{eqnarray*}
&    \Delta T =  E_T{(y_{\text{tar}}}) - E_T({y_{\text{ref}}}),~
    \Delta I =  E_I(\xb_{\text{gen}}) - E_I(\xb_{\text{ref}}).
 \end{eqnarray*}
Here, $E_I$ and $E_T$ are CLIP’s image and text encoders, respectively, and
 ${y_{\text{ref}}}, \xb_{\text{ref}}$ are the source domain text and image, respectively. The manipulated images guided by the directional CLIP loss are known robust to mode-collapse issues because by aligning the direction between the image representations with the direction between the reference text and the target text, distinct images should be generated. Also, it is more robust to adversarial attacks because the perturbation will be different depending on images \cite{radford2021learning}. More related works are illustrated in Supplementary Section \textred{A}.


\section{DiffusionCLIP}
\label{sec:method}

\begin{figure}[!htb]
    \centering
    \includegraphics[width=0.9\linewidth]{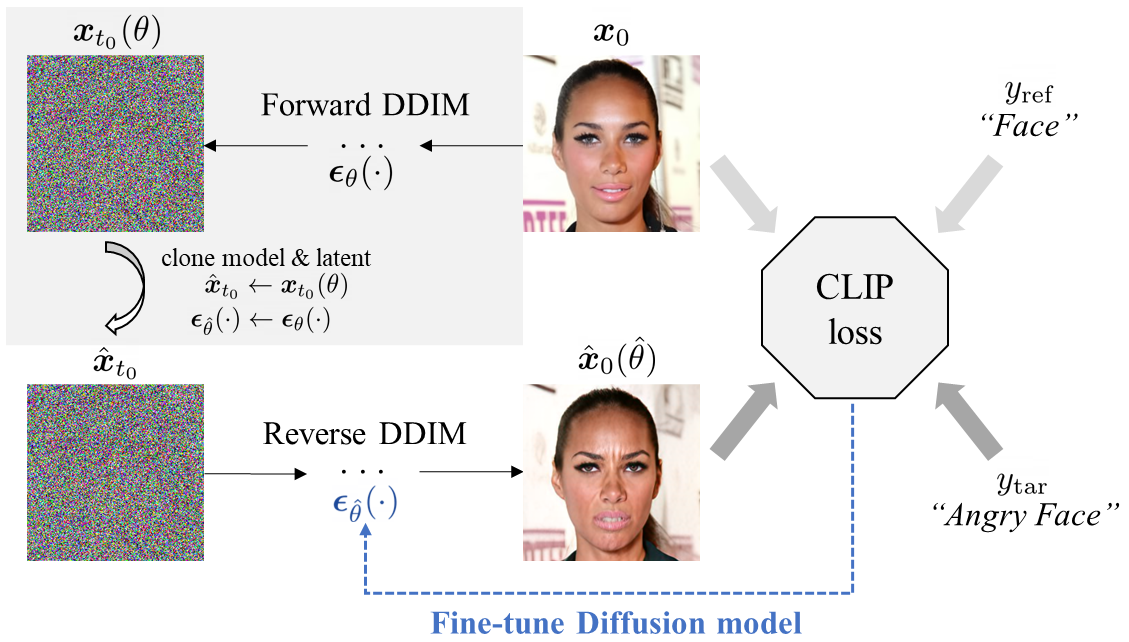}
    \vspace{-0.5em}
    \caption{Overview of DiffusionCLIP. The input image is first converted to the latent via diffusion models. Then, guided by directional CLIP loss, the diffusion model is fine-tuned, and the updated sample is generated during reverse diffusion.}
    \vspace{-0.5em}
    \label{fig:2_method}
\end{figure}

The overall flow of the proposed DiffusionCLIP for image manipulation is shown in Fig.~\ref{fig:2_method}.
Here, the input image $\xb_0$ is first converted to the latent { $\xb_{t_0}(\theta)$ using a  pretrained diffusion model $\epsilonb_\theta$.} Then, guided by the CLIP loss, the diffusion model at the reverse path is fine-tuned {to generate samples driven by the target text $y_\text{tar}$}. The deterministic forward-reverse processes are based on DDIM \cite{song2020denoising}. For translation between unseen domains, the latent generation is also done by forward DDPM \cite{ho2020denoising} process as will be explained later.

\subsection{DiffusionCLIP Fine-tuning}

In terms of fine-tuning, one could modify the latent or the diffusion model itself. We found that direct model fine-tuning is more effective, as analyzed in Supplementary Section \textred{D}.
Specifically, to fine-tune the reverse diffusion model $\epsilonb_{{\theta}}$,  we use the following objective composed of the directional CLIP loss $\mathcal{L}_\text{direction}$ and the identity loss $\mathcal{L}_\text{ID}$:
\begin{align}
    \label{eq:main_objective}
    {\mathcal{L}_{\text{direction}}\left(\hat\xb_0(\hat{\theta}),y_{\text{tar}}; \xb_{0},y_{\text{ref}}\right)+{\mathcal{L}_{\text{id}}( \hat\xb_0(\hat{\theta}), \xb_0) }},
    \vspace{-0.5em}
\end{align}
where $\xb_0$ is the original image, { $\hat\xb_0(\hat{\theta})$ is the generated image from the latent $\xb_{t_0}$ with the optimized parameter $\hat\theta$,}
 ${y_{\text{ref}}}$ is the reference text, ${y_{\text{tar}}}$ is the target text given for image manipulation.
 
 Here, the CLIP loss is the key component to supervise the optimization. 
 Of two types of CLIP losses as discussed above, we employ directional CLIP loss as a guidance thanks to the appealing properties as mentioned in Section~\ref{sec:CLIP}. For the text prompt, directional CLIP loss requires a reference text  ${y_{\text{ref}}}$ and a target text ${y_{\text{tar}}}$ while training. For example, in the case of changing the expression of a given face image into an angry expression, we can use `face' as a reference text and `angry face' as a target text. In this paper, we often use concise words to refer to each text prompt (e.g. `tanned face' to `tanned').
 
The identity loss $\mathcal{L}_{\text{id}}$ is employed to prevent the unwanted changes and preserve the identity of the object. We generally use $\ell_1$ loss as identity loss, and in case of human face image manipulation, face identity loss in \cite{deng2019arcface} is added:
\small
\begin{align}
{
 { \mathcal{L}_{\text{id}}( \hat\xb_0(\hat{\theta}), \xb_0 ) = \lambda_{\text{L1}} \|\xb_0 - \hat{\xb}_0( \hat{\theta}) \| + \lambda_\text{face}\mathcal{L}_{\text{face}}(\hat\xb_0(\hat{\theta}), \xb_0 )},}
\end{align}
\normalsize
{where $\Lc_{\text{face}}$ is the face identity loss \cite{deng2019arcface}, and $\lambda_{\text{L1}} \ge 0$ and $\lambda_\text{face} \ge 0 $ are weight parameters for each loss.} The necessity of identity losses depends on the types of the control. For some controls, the preservation of pixel similarity and the human identity are significant (e.g. expression, hair color) while others prefer the severe shape and color changes (e.g. artworks, change of species).

\begin{figure}[!htb]
    \centering
    \vspace{-0.5em}
    \includegraphics[width=0.9\linewidth]{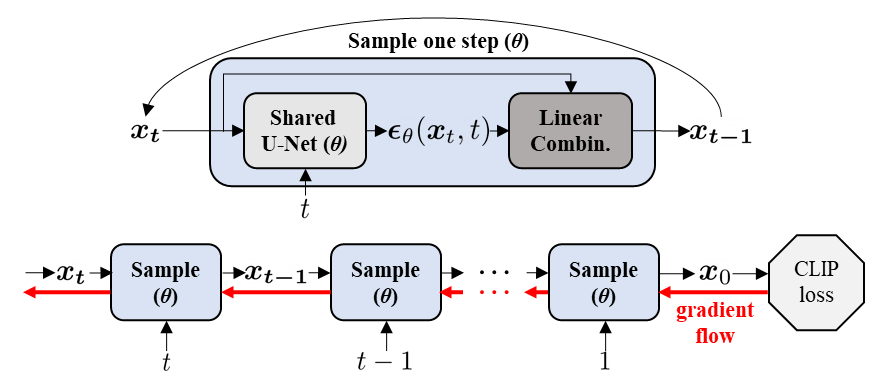}
    \vspace{-0.5em}
    \caption{Gradient flows during fine-tuning the diffusion model with the shared architecture across $t$.}
    \label{fig:3_shared}
     \vspace{-0.8em}
\end{figure}

Existing diffusion models \cite{ho2020denoising,song2020denoising, dhariwal2021diffusion} adopt the shared U-Net \cite{ronneberger2015u} architecture for all $t$, by inserting the information of $t$ using sinusoidal position embedding as used in the Transformer \cite{vaswani2017attention}. With this architecture, the gradient flow during DiffusionCLIP fine-tuning can be represented as Fig. \ref{fig:3_shared}, which is a similar process of training recursive neural network \cite{rumelhart1985learning}. 

Once the diffusion model is fine-tuned, any image from the pretrained domain can be manipulated into the image corresponding to the target text $y_\text{tar}$ as illustrated in Fig.~\ref{fig:4_applications}(a). {For details of the fine-tuning procedure and the model architecture, see Supplementary Section \textred{B} and \textred{C}.} 

\subsection{Forward Diffusion and Generative Process}
\label{sec:process}

As the DDPM sampling process in Eq.~\ref{eq:reverse_ddpm} is stochastic, the samples generated from the same latent will be different every time. 
Even if the sampling process is deterministic, the forward process of DDPM, where the random Gaussian noise is added as in Eq.~\ref{eq:forward_ddpm}, is also stochastic, hence the reconstruction of the original image is not guaranteed. To fully leverage the image synthesis performance of diffusion models with the purpose of image manipulation, we require the deterministic process both in the forward and reverse direction with pretrained diffusion models for successful image manipulation. On the other hand, for the image translation between unseen domains, stochastic sampling by DDPM is often helpful, which will be discussed in more detail later.

For the full inversion, we adopt deterministic reverse DDIM process \cite{song2020denoising, dhariwal2021diffusion} as generative process and ODE approximation of its reversal as a forward diffusion process. 
Specifically, {the deterministic 
forward DDIM process to obtain latent is represented as}: 
\begin{align}
\label{eq:forward_ddim}
\vx_{t+1} = \sqrt{\alpha_{t+1}}\vf_\theta(\vx_{t}, t) +  \sqrt{1 - \alpha_{t+1}}\bm{\epsilon}_{{\theta}}(\vx_{t}, t) 
\end{align}
and {the deterministic reverse DDIM process to generate sample from the obtained latent becomes}:
\begin{align}
\label{eq:reverse_ddim}
\xb_{t-1} = \sqrt{\alpha_{t-1}}\fb_\theta(\xb_{t}, t) +  \sqrt{1 - \alpha_{t-1}}\bm{\epsilon}_{\theta}(\xb_{t}, t) 
\end{align}
where $\vf_\theta$ is defined in Eq. \ref{eq:f}. {For the derivations of ODE approximation, see Supplementary Sec \textred{A}.}

Another important contribution of DiffusionCLIP is a fast sampling strategy.
Specifically, instead of performing forward diffusion until the last time step $T$,
we found that we can accelerate the forward diffusion by performing up to $t_0<T$, which we call `return step'.
 We can further accelerate training by using fewer discretization steps between $[1,t_0]$, denoted as $S_{\text{for}}$ and $S_{\text{gen}}$ for forward diffusion and generative process, respectively \cite{song2020denoising}. Through qualitative and quantitative analyses, we found the optimal groups of hyperparameters for $t_0, S_{\text{for}}$ and $S_{\text{gen}}$. For example, when $T$ is set to 1000 as a common choice \cite{ho2020denoising,song2020denoising, dhariwal2021diffusion}, 
the choices of $t_0 \in [300,600]$ and $(S_{\text{for}},S_{\text{gen}})=(40,6)$ satisfy our goal. Although $S_{\text{gen}}=6$ may give imperfect reconstruction, we found that the identity of the object that is required for training is preserved sufficiently. 
We will show the results of quantitative and qualitative analyses on $S_{\text{for}}, S_{\text{gen}}$ and $t_0$ later through experiments and Supplementary Section \textred{F}.

Lastly, if several latents have been precomputed (grey square region in Fig.~\ref{fig:2_method}), we can further reduce the time for fine-tuning by recycling the latent to synthesize other attributes. With these settings, the fine-tuning is finished in {$1{\sim}7$ minutes} on NVIDIA Quardro RTX 6000. 
\begin{figure}[!t]
    \centering
    \includegraphics[width=1.0\linewidth]{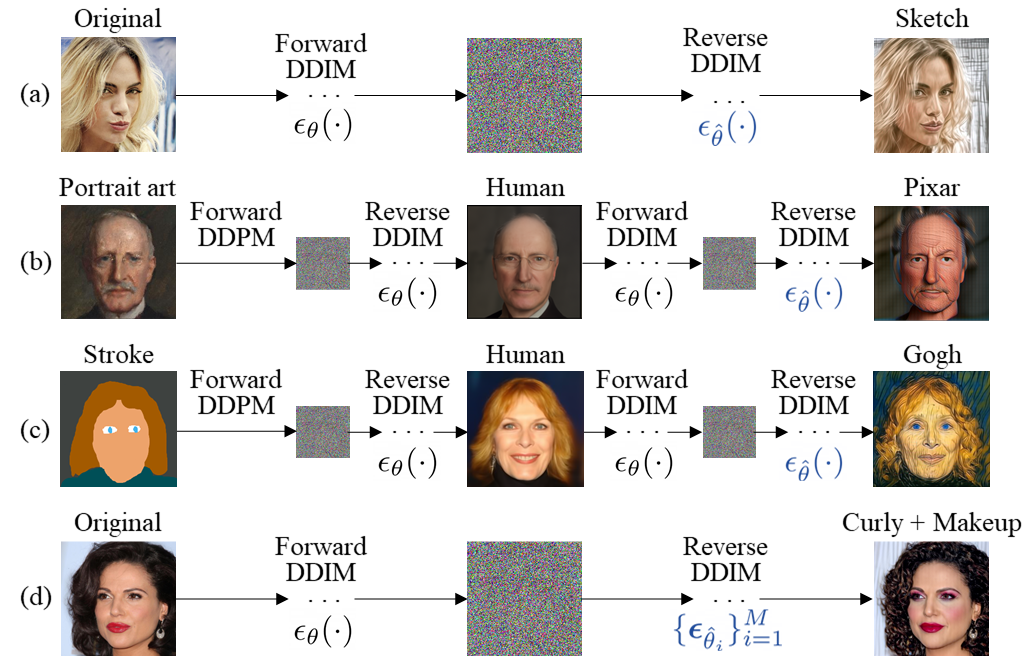}
    \caption{Novel applications of DiffusionCLIP. (a) Manipulation of images in pretrained domain to CLIP-guided domain. (b) Image translation between unseen domains. (c) Stroke-conditioned image generation in an unseen domain. (d) Multi-attribute transfer. $\epsilonb_\theta$ and $\epsilonb_{\hat{\theta}}$ indicate the original pretrained and fine-tuned diffusion models, respectively.}
    \vspace{-1em}
    \label{fig:4_applications}
\end{figure}

\begin{figure*}[!ht]
    \centering
    \includegraphics[width=0.95\linewidth]{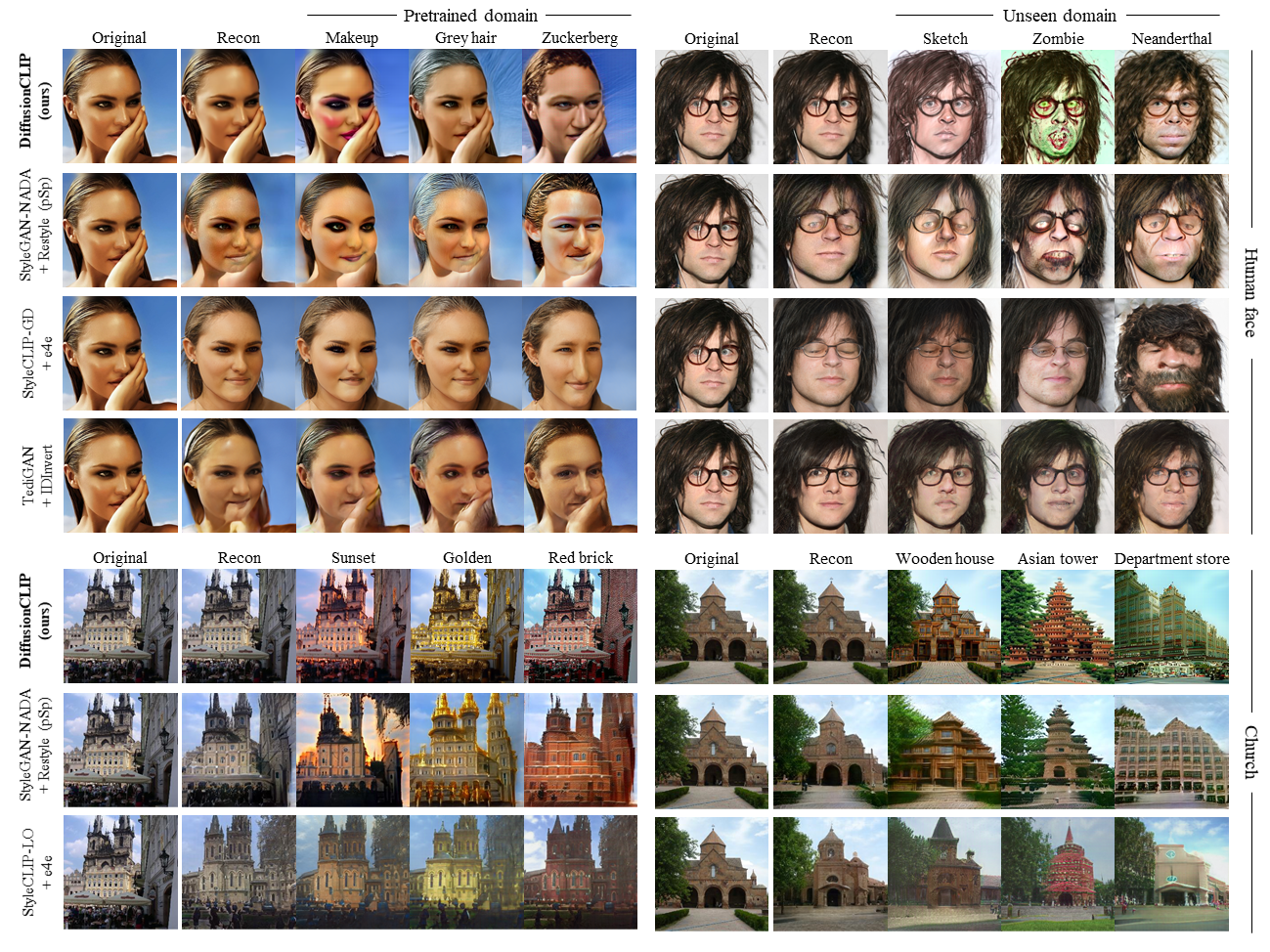}
   \vspace{-1.5em}
    \caption{Comparison with the state-of-the-art text-driven manipulation methods: TediGAN \cite{xia2021tedigan}, StyleCLIP \cite{patashnik2021styleclip} and StyleGAN-NADA \cite{gal2021stylegan}. StyleCLIP-LO and StyleCLIP-GD refer to the latent optimization (LO) and global direction (GD) methods of StyleCLIP.}
   \vspace{-0.5em}
    \label{fig:5_comparision}
\end{figure*}

\subsection{Image Translation between Unseen Domains}
The fine-tuned models through DiffusionCLIP can be leveraged to perform the additional novel image manipulation tasks as shown in Fig.~\ref{fig:4_applications}.

First, we can perform image translation from an unseen domain to another unseen domain, and stroke-conditioned image synthesis in an unseen domain as described in {Fig.~\ref{fig:4_applications}(b) and (c), respectively}. A key idea to address this difficult problem
is to bridge between two domains by inserting the diffusion models trained on the dataset that is relatively easy to collect. 
Specifically, in \cite{choi2021ilvr, meng2021sdedit}, it was found that with pretrained diffusion models, images trained from the unseen domain can be translated into the images in the trained domain. By combining this method with DiffsuionCLIP, we can now translate the images in zero-shot settings for both source and target domains. 
{Specifically, the images in the source unseen domain $\xb_0$ are first perturbed through the forward DDPM process in Eq. \ref{eq:forward_ddpm} until enough time step {$t_0$} when the domain-related component are blurred but the identity or semantics of object is preserved. This is usually set to 500. Next, the images in the pretrained domain $\xb'_{0}$ are sampled with the original pretrained model $\epsilonb_\theta$ using reverse DDIM process in Eq. \ref{eq:reverse_ddim}. 
Then, $\xb'_{0}$ is manipulated into the image $\hat{\xb_0}$ in the CLIP-guided unseen domain as we do in {Fig.~\ref{fig:4_applications}(a)} with the fine-tuned model $\epsilonb_{\hat{\theta}}$. }

\begin{figure*}[!hbt]
    \centering
    \includegraphics[width=0.95\linewidth]{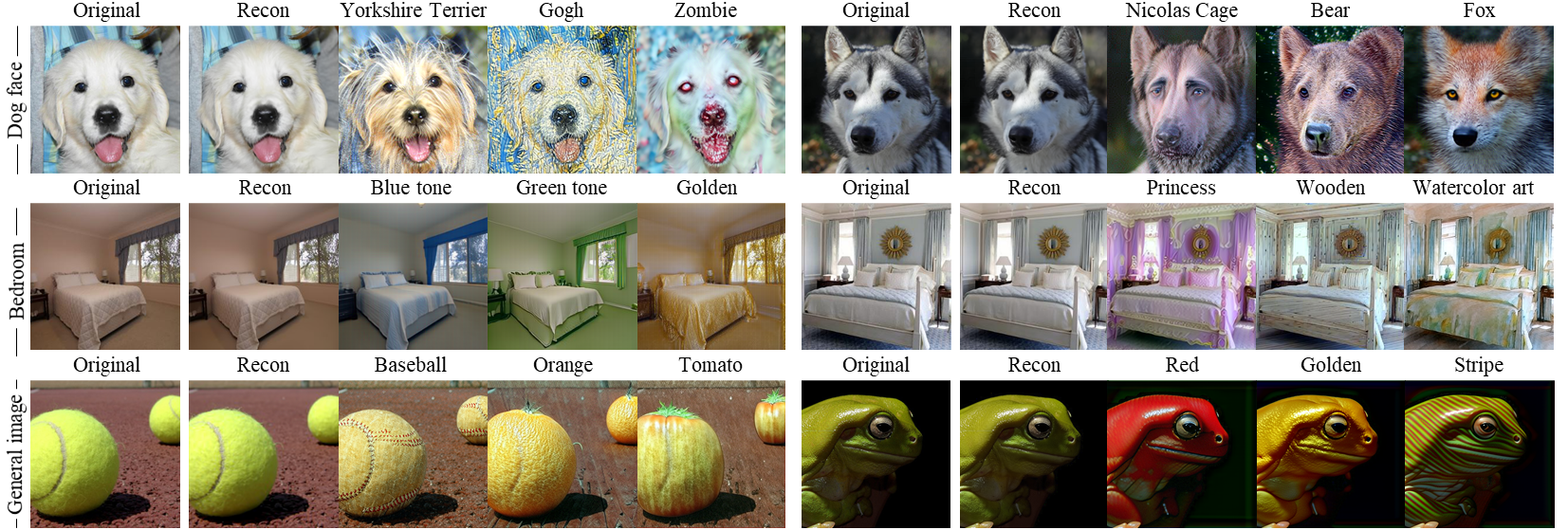}
    \vspace{-0.5em}
    \caption{\add{Manipulation results of real dog face, bedroom and general images using DiffusionCLIP.}}
    \vspace{-1em}
    \label{fig:6_more_results}
\end{figure*}

\subsection{Noise Combination}
\paragraph{Multi-attribute transfer.}
We discover that when the noises predicted from multiple fine-tuned models {$\{\epsilonb_{\hat{\theta}_i}\}^M_{i=1}$} are combined during the sampling, multiple attributes can be changed through only one sampling process as described in  Fig.~\ref{fig:4_applications}(d). {Therefore, we can flexibly mix several single attribute fine-tuned models with different combinations without having to fine-tune new models with target texts that define multiple attributes.} In detail, we first invert the image with the original pretrained diffusion model and use the multiple diffusion models by the following sampling rule:
\begin{equation}
\begin{split}
 {{\xb}}_{t-1} = &\sqrt{\alpha_{t-1}} \textstyle \sum\nolimits_{i=1}^M \gamma_i(t) \vf_{\hat{\theta}_i}(\boldsymbol{\xb}_t, t) \\
&+  \sqrt{1 - \alpha_{t-1}} \textstyle \sum\nolimits_{i=1}^M \gamma_i(t) \boldsymbol{\epsilon}_{\hat{\theta}_i}(\xb_t, t),
\end{split}
\end{equation}
\label{multi_attribute_transfer}
where  $\{\gamma_i(t)\}^T_{t=1}$ is the sequence of weights of each fine-tuned model $\epsilonb_{\hat{\theta}_i}$ satisfying  $\sum_{i=1}^M \gamma_i(t) = 1$ , which can be used for controlling the degree of each attribute. From  Eq. \ref{score_ddpm_relation}, we can interpret this sampling process as increasing the joint probability of conditional distributions as following:
\begin{equation}
\small
\textstyle \sum\nolimits_{i=1}^M \gamma_i(t) \boldsymbol{\epsilon}_{\hat{\theta}_i}(\xb_t, t) \propto -\nabla_{\xb_t}\log \prod\nolimits_{i=1}^Mp_{\hat{\theta}_i}(\xb_t | y_{\text{tar},i})^{\gamma_i(t)},
\normalsize
\end{equation}
where $y_\text{tar, i}$ is the target text for each fine-tuned model  $\epsilonb_{\hat{\theta}_i}$.

In the existing works \cite{choi2018stargan, choi2020stargan}, users require the combination of tricky task-specific loss designs or dataset preparation with large manual effort for the task, while ours enable the task in a natural way without such effort.

\paragraph{Continuous transition.} We can also apply the above noise combination method for controlling the degree of change during single attribute manipulation. By mixing the noise from the original pretrained model $\epsilonb_{{\theta}}$ and the fine-tuned model $\epsilonb_{\hat{\theta}}$ with respect to a degree of change $\gamma \in [0,1]$, we can perform interpolation between the original image and the manipulated image smoothly. 

For more details and pseudo-codes of the aforementioned applications, see Supplementary Section \textred{B}.

\section{Experiments}
\label{sec:experiment}
For all manipulation results by DiffusionCLIP, we use $256^2$ size of images. We used the models pretrained on CelebA-HQ \cite{karras2017progressive}, AFHQ-Dog \cite{choi2020starganv2}, LSUN-Bedroom and LSUN-Church \cite{yu15lsun} datasets for manipulating images of human faces, dogs, bedrooms, and churches, respectively. We use images from the testset of these datasets for the test.
To fine-tune diffusion models, we use Adam optmizer with an initial learning rate of 4e-6 which is increased linearly by 1.2 per 50 iterations. We set $\lambda_\text{L1}$ and $\lambda_\text{ID}$ to 0.3 and 0.3 if used. As mentioned in Section~\ref{sec:process}, we set $t_0$ in [300, 600] when the total timestep $T$ is 1000. We set $(S_{\text{for}},S_{\text{gen}})=(40,6)$ for training; and to $(200, 40)$ for the test time. Also, we precomputed the latents of 50 real images of size $256^2$ {in each training set of pretrained dataset}. {For more detailed hyperparameter settings, see Supplementary Section \textred{F}.}

\begin{table}[!htb]
\caption{Quantitative comparison for face image reconstruction.}\label{tab:compare_inversion}
\vspace{-0.5em}
\centering
\begin{adjustbox}{width=0.62\linewidth}
\begin{tabular}{lccc}
\toprule
Method & MAE ↓ & LPIPS ↓ & SSIM ↑ \\
 \midrule
Optimization & 0.061 & 0.126 & 0.875 \\
pSp & 0.079 & 0.169 & 0.793 \\
e4e & 0.092 & 0.221 & 0.742 \\
ReStyle w pSp & 0.073 & 0.145 & 0.823 \\
ReStyle w e4e & 0.089 & 0.202 & 0.758 \\
HFGI w e4e & 0.062 & 0.127 & 0.877 \\
 \midrule
\textbf{\add{Diffusion ($\bm{t_0=300}$)}} & \textbf{0.020} & \textbf{0.073} & \textbf{0.914} \\
\add{Diffusion} ($t_0=400$) & 0.021 & 0.076 & 0.910 \\
\add{Diffusion} ($t_0=500$) & 0.022 & 0.082 & 0.901 \\
\add{Diffusion} ($t_0=600$) & 0.024 & 0.087 & 0.893 \\
\bottomrule
\vspace{-2em}
\end{tabular}
\end{adjustbox}
\end{table}

\begin{table}[!htb]
\centering
\caption{Human evaluation results of real image manipulation on CelebA-HQ~\cite{karras2017progressive}. The reported values mean the preference rate of results from DiffusionCLIP against each method. }\label{tab:human_eval}%
\vspace{-0.5em}
\begin{adjustbox}{width=0.70\linewidth}
\begin{tabular}{cccc}
\toprule
 vs &   & \multicolumn{1}{c}{\begin{tabular}[c]{@{}c@{}}StyleGAN-NADA\\      (+ Restyle w pSp)\end{tabular}} & \multicolumn{1}{c}{\begin{tabular}[c]{@{}c@{}}StyleCLIP \\      (+ e4e)\end{tabular}} \\
  \midrule
\multirow{3}{*}{\begin{tabular}[c]{@{}c@{}}Hard\\ cases\end{tabular}} & In-domain & 69.85\% & 69.65\% \\
 & Out-of-domain & 79.60\% & 94.60\% \\
 & {All domains} & {73.10\%} & {77.97\%} \\
  \midrule
\multirow{3}{*}{\begin{tabular}[c]{@{}c@{}}General\\ cases\end{tabular}} & In-domain & 58.05\% & 50.10\% \\
 & Out-of-domain & 71.03\% & 88.90\% \\
 & {All domains} & {62.47\%} & {63.03\%} \\
\bottomrule
\vspace{-2em}
\end{tabular}
\end{adjustbox}
\end{table}

\begin{table}[!htb]
\centering
\caption{{Quantitative evaluation results. Our goal is to achieve the better score in terms of Directional CLIP similarity ($\mathbf{\mathcal{S}_\text{dir}}$), segmentation-consistency (SC), and face identity similarity (ID).}}\label{tab:non_subjective}%
\vspace{-0.5em}
\begin{adjustbox}{width=0.72\linewidth}
\begin{tabular}{lccccc}
                    \toprule
                    & \multicolumn{3}{c}{\small{CelebA-HQ}} & \multicolumn{2}{c}{\small{LSUN-Church}} \\ 
                     \cmidrule(r){2-4} \cmidrule(l){5-6}
                     &  $\mathcal{S}_\text{dir}$↑ & SC↑ & ID↑  & $\mathcal{S}_\text{dir}$↑ & SC↑ \\
                     \midrule
                    StyleCLIP & 0.13 & 86.8\% & 0.35 & 0.13 & 67.9\% \\
                    StyleGAN-NADA & 0.16 & 89.4\% & 0.42 & 0.15 & 73.2\% \\
                     \midrule
                    \textbf{DiffusionCLIP (Ours)} & \textbf{0.17} & \textbf{93.7\%} & \textbf{0.70} & \textbf{0.20} & \textbf{78.1\%} \\
                    \bottomrule
\end{tabular}
\end{adjustbox}
\end{table}


\begin{figure}[!htb]
    \centering
    \includegraphics[width=0.9\linewidth]{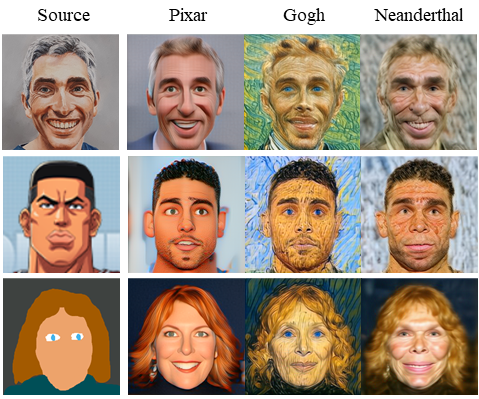}
    \vspace{-0.5em}
    \caption{Results of image translation between unseen domains.}
    \vspace{-0.5em}
    \label{fig:7_u2u_s2u}
\end{figure}

\begin{figure}[!htb]
    \centering
    \includegraphics[width=0.9\linewidth]{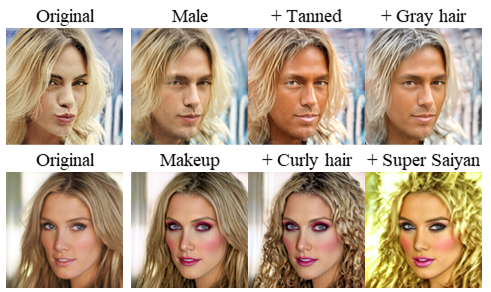}
    \vspace{-0.5em}
    \caption{Results of multi-attribute transfer.}
    \vspace{-1em}
    \label{fig:8_multi_attribute}
\end{figure}

\begin{figure}[!htb]
    \centering
    \includegraphics[width=0.9\linewidth]{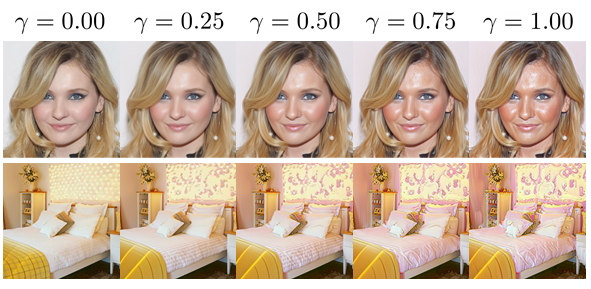}
    \vspace{-0.5em}
    \caption{Results of continuous transition.}
    \vspace{-1.5em}
    \label{fig:9_continuous_transition}
\end{figure}

\begin{figure}[!htb]
    \centering
    \includegraphics[width=0.9\linewidth]{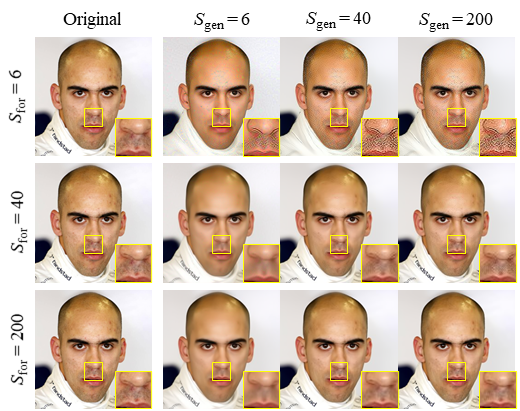}
    \vspace{-0.5em}
    \caption{Reconstruction results varying the number of forward diffusion steps $S_{\text{for}}$ and generative steps $S_\text{gen}$.}
    \vspace{-0.5em}
    \label{fig:10_varying_steps}
\end{figure}

\begin{figure}[!htb]
    \centering
    \vspace{-0.5em}
    \includegraphics[width=0.9\linewidth]{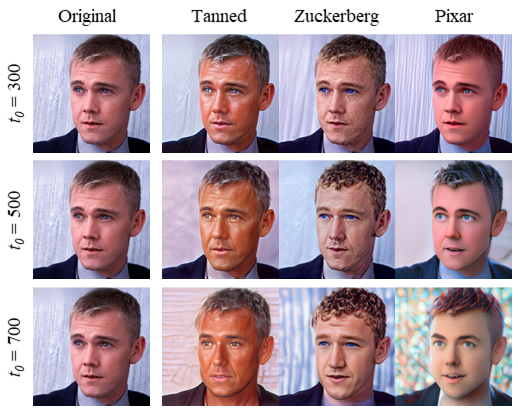}
    \vspace{-0.5em}
    \caption{Manipulation results depending on $t_0$ values.}
    \vspace{-1.5em}
    \label{fig:11_varying_t0}
\end{figure}
 
\subsection{Comparison and Evaluation}
\label{sec:comparsion}
\paragraph{Reconstruction.}To demonstrate the nearly perfect reconstruction performance of our method, we perform the quantitative comparison with SOTA GAN inversion methods, pSp \cite{richardson2021encoding}, e4e \cite{tov2021designing}, ReStyle \cite{alaluf2021restyle} and HFGI \cite{wang2021high}. As in Tab.~\ref{tab:compare_inversion}, our method shows higher reconstruction quality than all baselines in terms of all metrics: MAE, SSIM and LPIPS \cite{zhang2018perceptual}.

\paragraph{Qualitative comparison.}For the qualitative comparison of manipulation performance with other methods, we use the state-of-the-art text manipulation methods, TediGAN \cite{xia2021tedigan}, StyleCLIP \cite{patashnik2021styleclip} and StyleGAN-NADA \cite{gal2021stylegan} where images for the target control is not required similar to our method. StyleGAN2\cite{karras2020analyzing} pretrained on FFHQ-1024~\cite{karras2019style} and LSUN-Church-256~\cite{yu15lsun} is used for StyleCLIP and StyleGAN-NADA. StyleGAN \cite{karras2019style} pretrained on FFHQ-256~\cite{karras2019style} is used for TediGAN. For GAN inversion, e4e encoder~\cite{tov2021designing} is used for StyleCLIP latent optimization (LO) and global direction (GD), Restyle encoder~\cite{alaluf2021restyle} with pSp~\cite{richardson2021encoding} is used for StyleGAN-NADA, and IDInvert \cite{zhu2020domain} is used for TediGAN, as in their original papers. Face alignment algorithm is used for StyleCLIP and StyleGAN-NADA as their official implementations. Our method uses DDPM pretrained on CelebA-HQ-256~\cite{karras2017progressive} and LSUN-Church-256~\cite{yu15lsun}. 

As shown in Fig. \ref{fig:5_comparision}, SOTA GAN inversion methods fail to manipulate face images with novel poses and details producing distorted results. Furthermore, in the case of church images, the manipulation results can be recognized as the results from different buildings.
These results imply significant practical limitations. On the contrary, our reconstruction results are almost perfect even with fine details and background, which enables faithful manipulation. In addition to the manipulation in the pretrained domain, DiffusonCLIP can perform the manipulation into the unseen domain successfully, while StyleCLIP and TediGAN fail.

\paragraph{User study.}
We conduct user study to evaluate real face image manipulation performance on CelebA-HQ~\cite{karras2017progressive} with our method, StyleCLIP-GD \cite{patashnik2021styleclip} and StyleGAN-NADA \cite{gal2021stylegan}. We get 6000 votes from 50 people using a survey platform. {We use the first 20 images in CelebA-HQ testset as general cases and use another 20 images with novel views, hand pose, and fine details as hard cases.} 
For a fair comparison, we use 4 in-domain attributes (angry, makeup, beard, tanned) and 2 out-of-domain attributes (zombie, sketch), which are used in the studies of baselines. Here, we use official pretrained checkpoints and implementation for each approach.
 As shown in Tab.~\ref{tab:human_eval}, for both general cases and hard cases, all of the results from DiffusionCLIP are preferred compared to baselines ($> 50\%$). Of note, in hard cases, the preference rates for ours were all increased, demonstrating robust manipulation performance. It is remarkable that the high preference rates ($\approx 90\%$) against StyleCLIP in out-of-domain manipulation results suggest
 that our method significantly outperforms StyleCLIP in out-of-domain manipulation.

\paragraph{Quantitative evaluation.} {We also compare the manipulation performance using the following quality metrics: Directional CLIP similarity ($\mathbf{\mathcal{S}_\text{dir}}$), segmentation-consistency (SC),
and  face identity similarity (ID). To compute each metric, we use a pretrained CLIP~\cite{radford2021learning}, segmentation~\cite{yu2018bisenet,zhou2017scene, zhou2018semantic} and face recognition models~\cite{deng2019arcface}, respectively.
Then, during the translation between three attributes in CelebA-HQ (makeup, tanned, gray hair)~\cite{karras2017progressive} and  LSUN-Church (golden, red brick, sunset)~\cite{yu15lsun},
our goal is to achieve the better score in terms of $\mathbf{\mathcal{S}_\text{dir}}$, SC, and ID. As shown in Tab.~\ref{tab:non_subjective}, our method outperforms baselines in all metrics, demonstrating high attribute-correspondence ($\mathcal{S}_\text{dir}$) as well as well-preservation of identities without unintended changes (SC, ID).}

For more experimental details and results of the comparison, see Supplementary Section \textred{D} and \textred{E}.
\subsection{More Manipulation Results on Other Datasets}
\label{sec:more_result}
Fig. \ref{fig:6_more_results} presents more examples of image manipulations on dog face, bedroom and general images using the diffusion models pretrained on AFHQ-Dog-256 \cite{choi2020starganv2}, LSUN-Bedroom-256~\cite{yu15lsun} {and ImageNet-512~\cite{ILSVRC15} datasets, respectively. The results demonstrate that the reconstruction is nearly flawless and high-resolution images can be flexibly manipulated beyond the boundary of the trained domains. Especially, due to the diversity of the images in ImageNet, GAN-based inversion and its manipulation in the latent space of ImageNet show limited performance~\cite{daras2020your, bora2017compressed}. DiffusionCLIP enables the zero-shot text-driven manipulation of general images, moving a step forward to the general text-driven manipulation. For more results, see Supplementary Section \textred{E}.}

\subsection{Image Translation {between} Unseen Domains}
\label{sec:u2u_s2u_result}
With the fine-tuned diffusion models using DiffusionCLIP, we can even translate the images in one unseen domain to another unseen domain. Here, we are not required to collect the images in the source and target domains or introduce external models. In Fig.~\ref{fig:7_u2u_s2u}, we perform the image translation results from the portrait artworks and animation images to other unseen domains, Pixar, paintings by Gogh and Neanderthal men. We also show the successful image generation in the unseen domains from the stroke which is the rough image painting with several color blocks. These applications will be useful when enough images for both source and target domains are difficult to collect. 

\subsection{Noise Combination}
As shown in Fig.~\ref{fig:8_multi_attribute} we can change multiple attributes in one sampling. 
As discussed before, to perform the multi-attribute transfer, complex loss designs, as well as specific data collection with large manual efforts, aren't required.
Finally,  Fig.~\ref{fig:9_continuous_transition} shows that we can control the degree of change of single target attributes according to $\gamma$ by mixing noises from the original model and the fine-tuned model.





\subsection{Dependency on Hyperparameters}
 In Fig.~\ref{fig:10_varying_steps}, we show the results of the reconstruction performance depending on $S_{\text{for}}$, $S_{\text{gen}}$ when $t_0=500$. Even with $S_{\text{for}}=6$, we can see that the reconstruction preserves the identity well. When $S_{\text{for}}=40$, the result of $S_{\text{gen}}=6$ lose some high frequency details, but it's not the degree of ruining the training.
{When $S_{\text{for}}=200$ and $S_{\text{gen}}=40$, the reconstruction results are so excellent that we cannot differentiate the reconstruction with the result when the original images. Therefore, we just use $(S_{\text{for}}$, $S_{\text{gen}})=(40,6)$ for the training and $(S_{\text{for}}, S_{\text{gen}})=(200,40)$ for the inference.} 

We also show the results of manipulation by changing $t_0$ while fixing other parameters in Fig.~\ref{fig:11_varying_t0}. In case of skin color changes, 300 is enough. However, in case of the changes with severe shape changes such as the Pixar requires stepping back more as $t_0 = 500$ or $t_0 = 700$. Accordingly, we set different $t_0$ depending on the attributes. The additional analyses on hyperparameters and ablation studies are provided in Supplementary Section \textred{F}.

\section{Discussion and Conclusion}
In this paper, we proposed DiffusionCLIP, a method of text-guided image manipulation method using the pretrained diffusion models and CLIP loss. Thanks to the near-perfect inversion property, DiffusionCLIP has shown excellent performance for both in-domain and out-of-domain manipulation by fine-tuning diffusion models. We also presented several novel applications of using fine-tuned models by combining various sampling strategies.

There are limitations and societal risks on DiffusionCLIP. 
 Therefore, we advise users to make use of our method carefully for proper purposes. Further details on limitations and negative social impacts are given in Supplementary Section \textred{G} and \textred{H}.

{\small
\bibliographystyle{ieee_fullname}
\bibliography{egbib}
}

\clearpage
\appendix

\twocolumn[{%
\renewcommand\twocolumn[1][]{#1}%
\maketitle
\begin{center}
    \centering
    \captionsetup{type=figure}
    \vspace{-1em}
    \includegraphics[width=1.0\linewidth]{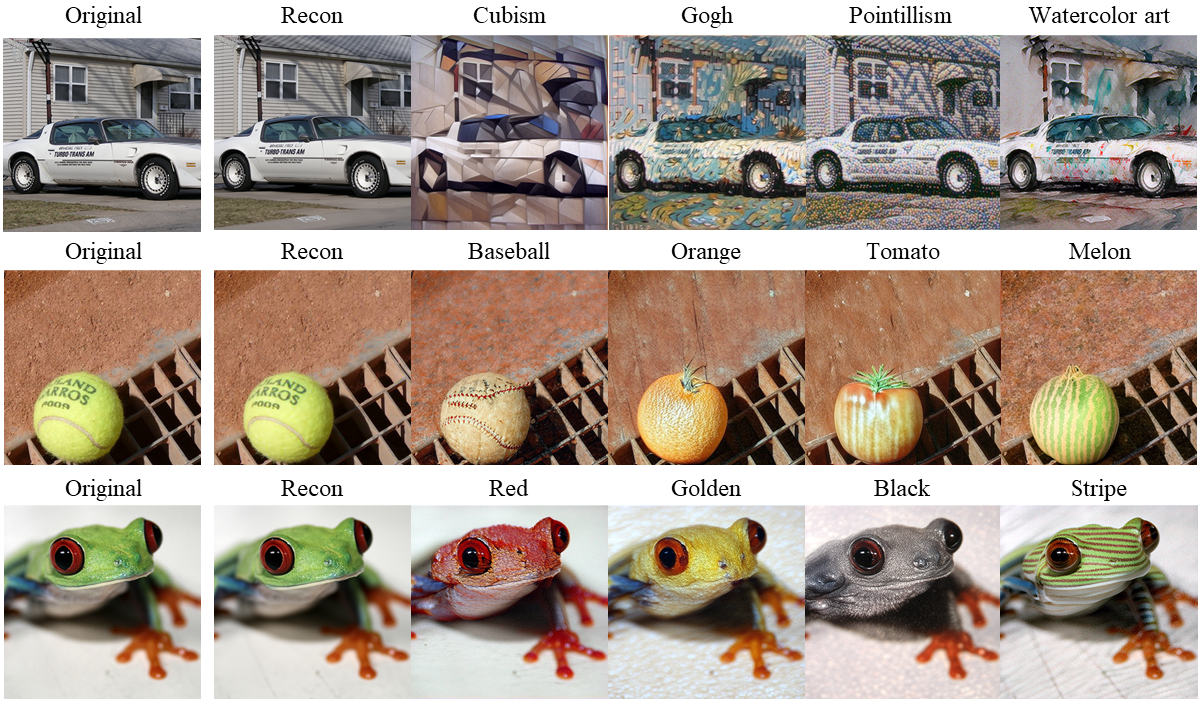}
    \captionof{figure}{DiffusionCLIP can even perform manipulation of $512\times512$ images using the ImageNet~~\cite{ILSVRC15} pretrained diffusion models.  Thanks to the near-perfect inversion capability, DiffusionCLIP  enables the zero-shot text-driven manipulation, moving a step forward to the general text-driven manipulation.  In contrast, due to the diversity of the images in ImageNet, GAN-based inversion  and its manipulation in the latent space of ImageNet  shows limited performance~\cite{daras2020your, bora2017compressed}. Hence, zero-shot text-driven manipulation using ImageNet pretrained GAN have been rarely explored. For more results, see Fig.~\ref{fig_s:comparision_vqgan}, \ref{fig_s:imagenet_general}, \ref{fig_s:imagenet_tennis} and \ref{fig_s:imagenet_frog}. }
    \label{fig_s:preview_imagenet}
\end{center}%
}]

    
\noindent\large\textbf{Supplementary Material}
\normalsize
\section{Details on Related Works}
\label{app:realted_works}
\subsection{DDPM, DDIM and ODE Approximation}
\paragraph{Denoising diffusion probabalistic models (DDPM).}

Diffusion probabilistic models \cite{ho2020denoising} are a class of latent variable models based on  forward  and  reverse processes. Suppose that our model distribution  $p_\theta(\xb_0)$  tries to approximate a data distribution $q(\xb_0)$. Let $\mathcal{X}$ denote the sample space for $\xb_0$ generated from a sequence of latent variables $\xb_t$ for $t = 1, \cdots , T$, where $\xb_T \sim \mathcal{N}(\mathbf{0,I})$. In the forward process, noises are gradually added to data $\xb_0$ and the latent  sequence set $\xb_{1:T}$ are generated through the following Markov chain upon a variance schedule defined by  $\{\beta_t\}^T_{t=1}$:
\begin{align}
    q(\xb_{1:T}) :=	\prod_{t=1}^T q(\xb_{t}|\xb_{t-1}),  
\end{align}
where
\begin{align}
    q(\xb_{t}|\xb_{t-1}) := \mathcal{N}(\sqrt{1-\beta_t} \xb_{t-1}, \beta_t\mathbf{I} ).
\end{align}
Then, $q(\xb_{t}|\xb_{0})$ can be represented in a closed form as $q(\xb_{t}|\xb_{0}) = \mathcal{N}(\xb_{t};\sqrt{\alpha_t} \xb_{0}, (1-\alpha_t)\mathbf{I})$, where $\alpha_t := 1-\beta_t$ and $\bar\alpha_t:= \prod_{s=1}^{t} {(1-\beta_s)}$. Then, we can sample $\xb_t$ as:
 \begin{align}
    \xb_t = \sqrt{\bar\alpha_t}\xb_0  + \sqrt{1 - \bar\alpha_t}\wb, \  \text{where} \ \wb \sim \mathcal{N} (\mathbf{0,I}).
     \label{eq:sample_xt}
\end{align}
 In the reverse process, $\xb_T$ is denoised to generate $\xb_0$ through the following Markov process:
\begin{align}
    p_\theta(\xb_{0:T}) := p(\xb_T)	\prod_{t=1}^T p_\theta(\xb_{t-1}|\xb_{t}), 
\end{align}
where  $\xb_T \sim \mathcal{N} (\mathbf{0,I})$ and
\begin{align}
    p_\theta(\xb_{t-1}|\xb_{t}) := \mathcal{N}(\mub_{\theta}(\xb_t, t), \boldsymbol{\Sigma}_{\theta}(\xb_t, t)\mathbf{I}),
\end{align}
where  $\boldsymbol{\Sigma}_{\theta}(\xb_t, t)$ is set to be learnable to improve the sample quality \cite{nichol2021improved}
and
   \begin{align}
    \mub_\theta(\xb_t, t) = \frac{1}{\sqrt{\alpha_t}}\left(\xb_t - \frac{1-\alpha_t}{\sqrt{1-\bar\alpha_t} }\epsilonb_\theta(\xb_t, t)\right).
\end{align}
and
the neural network $\epsilonb_\theta(\xb_t, t)$  is trained with the following improved objective \cite{ho2020denoising}:
 \begin{align}
 \small
    \mathcal{L}_\text{simple} := \mathbb{E}_{\xb_0, \bm{\wb}, t} ||\bm{\wb}-\epsilonb_\theta( \sqrt{\bar\alpha_t}\xb_0  + \sqrt{1 - \bar\alpha_t}\wb, t) ||^2_2,
    \label{eq:L_simple}
\normalsize
\end{align}
where $\wb \sim \mathcal{N} (\mathbf{0,I})$. 
%
%
%

\paragraph{Denoising diffusion implicit models (DDIM).}    
An alternative non-Markovian forward process that has the same forward marginals as DDPM and corresponding sampling process is proposed in \cite{song2020denoising}.
Here, the forward diffusion is described by
%
%
$${\xb_t = \sqrt{\bar\alpha_t}\xb_0 +\sqrt{1-\bar\alpha_t}\zb},$$
while the reverse diffusion
can be represented as following:
\small
\begin{align}
\xb_{t-1} = \sqrt{\bar\alpha_{t-1}}\fb_\theta(\xb_{t}, t) +  \sqrt{1 - \bar\alpha_{t-1} - \sigma^2_t}{\epsilonb}_{\theta}(\xb_{t}, t) + \sigma^2_t\zb,
\label{eq:ddim2}
\end{align}
\normalsize
where $\bm{\zb} \sim \mathcal{N} (\mathbf{0,I})$ 
and $\vf_\theta(\xb_{t}, t)$ is a the prediction of $\xb_0$ at $t$ given $\xb_t$ :
\begin{align}\label{eq:f}
  \vf_\theta(\xb_{t}, t):= \frac{\xb_{t} - \sqrt{1-\bar\alpha_t}\epsilonb_{\theta}(\xb_t,t)}{\sqrt{\bar\alpha_{t}}},
 \end{align}
 and $\epsilonb_\theta(\vx_t, t)$  is computed by  (\ref{eq:L_simple}).
 
This sampling allows using different reverse samplers by changing the variance of the reverse noise $\sigma_t$. Especially, by setting this noise to 0, which is a DDIM sampling process \cite{song2020denoising}, the sampling process becomes deterministic, enabling to conversation latent variables into the data consistently and to sample with fewer steps. 

%
%
%
 \paragraph{ODE approximation.}
 In fact, DDIM can be considered as a Euler method to solve ODE.
 Specifically,  Eq.~(\ref{eq:ddim2}) can be represented as:
\small
\begin{align}
\label{eq:euler_gen}
\sqrt{\frac{1}{\bar\alpha_{t-1}}}\xb_{t-1}  - \sqrt{\frac{1}{\bar\alpha_{t}}}\xb_{t}  = \left(\sqrt{\frac{1}{\bar\alpha_{t-1}}-1} - \sqrt{\frac{1}{\bar\alpha_{t}}-1}\right) \epsilonb_\theta(\xb_t, t)
\end{align}
\normalsize
If we set $\y_t := \sqrt{1/\bar\alpha_{t}}\xb_{t}$ and $p_t := \sqrt{1/\bar\alpha_{t}-1}$, we can rewrite Eq. (\ref{eq:euler_gen}) as follows:
\begin{align}
\y_{t-1} - \y_{t} = (p_{t-1} - p_t)\epsilonb_\theta(\xb_t, t).
\end{align}
In the limit of small steps, this equation goes to ODE 
$$d y_t =  \epsilonb_\theta(\xb_t, t) d p_t.$$
Then, the reversal of this ODE can be derived as follows:
\begin{align}
 \y_{t+1} - \y_{t} = (p_{t+1} - p_t)\epsilonb_\theta(\xb_t, t),
 \end{align}
which becomes:
 \small
 \begin{align}
 \sqrt{\frac{1}{\bar\alpha_{t+1}}}\xb_{t+1}  - \sqrt{\frac{1}{\bar\alpha_{t}}}\xb_{t}  = \left(\sqrt{\frac{1}{\bar\alpha_{t+1}}-1} - \sqrt{\frac{1}{\bar\alpha_{t}}-1}\right) \epsilonb_\theta(\xb_t, t).
 \end{align}
 \normalsize
 Finally, the above equation can be written as:
 \begin{align}
 \vx_{t+1} = \sqrt{\bar\alpha_{t+1}}\vf_\theta(\vx_{t}, t) +  \sqrt{1 - \bar\alpha_{t+1}}\bm{\epsilon}_{{\theta}}(\vx_{t}, t) ,
\end{align}
which is equal to our forward DDIM process formulation that is used in Sec. 3.2.

\subsection{Additional Related Works}
\paragraph{{Diffusion}-based image manipulation.}
Recent diffusion models have demonstrated impressive performance in image generation \cite{sohl2015deep, song2019generative,ho2020denoising,song2020denoising,song2020score,jolicoeur2020adversarial, dhariwal2021diffusion}
with additional advantages of great mode coverage and stable training. 

Despite this recent progress, only a few studies \cite{choi2021ilvr, meng2021sdedit} have been carried out for image manipulation with diffusion models, such as local editing and the image translation from unseen domain to the trained domain. In ILVR \cite{choi2021ilvr}, image translation where the low-frequency component of the reference image is conditioned at each transition during the sampling process is introduced. In SDEdit \cite{meng2021sdedit}, images with the user’s local edit or strokes are first noised via the stochastic SDE process, and subsequently denoised by simulating the reverse SDE to generate the realistic image in the pretrained domain. However, it is not clear how these methods can be extended for more general image manipulation applications, such as attribute manipulation, translation from the trained domain to multiple unseen domains, etc.

On the other hand, DiffusionCLIP enables text-guided image manipulation with an infinite number of types of text-driven attributes, and translation of images in the pretrained or an unseen domain to another unseen domain.

\paragraph{GAN-based image manipulation.}

Image manipulation methods have been mostly implemented using GAN models. Conditional GAN methods \cite{isola2017image, zhu2020sean, wang2018high,zhu2017unpaired, portenier2018faceshop, park2019semantic, chen2017photographic, dekel2018sparse} learn direct mappings from original images to target images. However, these methods need additional training and collection of the dataset with a huge amount of manual effort whenever the new controls are necessary. 

In GAN inversion based methods \cite{zhu2016generative,brock2016neural, richardson2021encoding, tov2021designing, bau2020semantic, roich2021pivotal, abdal2019image2stylegan, abdal2020image2stylegan++, gu2020image, wu2021stylespace, alaluf2021restyle, wang2021high, zhu2020domain},
an input image is first converted to a latent vector so that the image can be manipulated by modifying the latent or fine-tuning the generator. In recent works \cite{patashnik2021styleclip, gal2021stylegan}, GAN inversion is combined with the CLIP loss \cite{radford2021learning}, so that image manipulation  given simple text prompts can be achieved without additional training dataset for target distribution. 

However, image manipulation by GAN inversion still demands further investigation,
because many datasets are still hard to invert due to 
the limited inversion capability of GAN models  \cite{richardson2021encoding, karras2019style, huh2020transforming}.
Even the encoder-based GAN inversion approaches \cite{tov2021designing, richardson2021encoding, alaluf2021restyle}, which is the current state-of-the-art (SOTA) methods, often fail to reconstruct images with novel poses, views, and details, inducing the unintended change in the manipulation results. This issue becomes even worse in the case of images from a dataset with high variances such as church images in LSUN Church~\cite{yu15lsun} or ImageNet dataset~\cite{ILSVRC15}.

{On the other hand, DiffusionCLIP allows near-perfect inversions, so that it can perform zero-shot text-driven image manipulation successfully, preserving important details even for images from a dataset with high variance. We can even translate the image from an unseen domain into another unseen domain or generate images in an unseen domain from the strokes.}
In the following, we illustrate the detailed procedure with pseudocode.

\section{Details on Methods}
\label{app:method}


\subsection{DiffusionCLIP Fine-tuning}
\label{app:method_fine}
We adopt a two-step approach as detailed in Algorithm~\ref{algo:fine-tuning}. First,  real images or sampled images from pretrained domain  $\{ \xb^{(i)}_0 \}^N_{i=1}$ are inverted to the latents $\{ \xb^{(i)}_{t_0} \}^N_{i=1}$ via deterministic forward DDIM processes \cite{song2020denoising} with the pretrained diffusion model $\epsilonb_\theta$. 
To accelerate the process, instead of performing forward diffusion until the last time step $T$,
{we  use fewer discretization steps $\{ \tau_s\}^{S_{\textnormal{for}}}_{s=1}$ such that $\tau_1=0,\tau_{  \scalebox{.8}{$\scriptscriptstyle S_{\textnormal{for}} $}   }=t_0$.}

In the second step, we start to update $\epsilonb_{\hat{\theta}}$, a copy of the pretrained diffusion model. For each latent in $\{ \xb^{(i)}_{t_0} \}^N_{i=1}$, the image is sampled through the reverse DDIM process \cite{song2020denoising}  and the model is updated guided by CLIP loss $\mathcal{L}_\text{direction}$ and identity loss $\mathcal{L}_\text{ID}$ to generate images that represent $y_\text{tar}$. The second step is repeated $K$ times until converged.

 \begin{algorithm}[h!]
    \caption{DiffuisonCLIP fine-tuning\label{algo:fine-tuning}}
    \DontPrintSemicolon
    \SetAlgoNoLine
    \SetAlgoVlined
    \SetKwProg{Fn}{Require}{:}{}
    \KwIn{$\epsilonb_{{\theta}}$ (pretrained model), $\{ \xb^{(i)}_0 \}^N_{i=1}$ (images to precompute), $y_\text{ref}$ (reference text), $y_\text{tar}$ (target text), $t_0$ (return step), $S_{\textnormal{for}}$ (\# of inversion steps),  $S_{\textnormal{gen}}$(\# of generation steps), $K$ (\# of fine-tuning iterations)} 
    \KwOut{$\epsilonb_{\hat{\theta}}$ (fine-tuned model)}
    \tcp{Step 1: Precompute latents}
    Define $\{ \tau_s\}^{S_{\textnormal{for}}}_{s=1}$ s.t. $\tau_1=0,\tau_{  \scalebox{.8}{$\scriptscriptstyle S_{\textnormal{for}} $}   }=t_0$. \;
    \For{$i = 1,2,\ldots, N$}{ 
        \For{$\ s = 1,2,\ldots, S_{\textnormal{for}}-1$}{
            ${\epsilonb \leftarrow \epsilonb_{{\theta}}(\xb^{(i)}_{\tau_{s}}, \tau_{s}) ; \ \  \fb \leftarrow \fb_\theta(\xb^{(i)}_{\tau_{s}},\tau_{s})}$ \;
            ${\xb^{(i)}_{\tau_{s+1}} \leftarrow \sqrt{\alpha_{\tau_{s+1}}}\fb + \sqrt{1 - \alpha_{\tau_{s+1}}}\epsilonb}$ \;
            }
        Save the latent $\xb^{(i)}_{t_0} = \xb^{(i)}_{ \tau_{  \scalebox{.8}{$\scriptscriptstyle S_{\textnormal{for}} $}   } }$. 
        
    }
    \tcp{Step 2: Update the diffusion model}
    Clone the pretrained model $\epsilonb_{\hat{\theta}} \leftarrow \epsilonb_{{\theta}}$\;
    Define $\{ \tau_s\}^{S_{\textnormal{gen}}}_{s=1}$ s.t. $\tau_1=0,\tau_{  \scalebox{.8}{$\scriptscriptstyle S_{\textnormal{gen}} $}   }=t_0$. \;
    \For{$k = 1,2,\ldots, K$}{
        \For{$i = 1,2,\ldots, N$}{
            Clone the latent $\hat{\xb}^{(i)}_{t_0} \leftarrow \xb^{(i)}_{t_0}$. \;
            \For{$\ s = S_{\textnormal{gen}}, S_{\textnormal{gen}}-1,\ldots, 2$}{
            ${\epsilonb \leftarrow \epsilonb_{\hat{\theta}}(\hat{\xb}^{(i)}_{\tau_{s}}, \tau_{s}) ; \ \  \fb \leftarrow \fb_{\hat{\theta}}(\hat{\xb}^{(i)}_{\tau_{s}},\tau_{s})}$ \;
            ${\hat{\xb}^{(i)}_{\tau_{s-1}} \leftarrow \sqrt{\alpha_{\tau_{s-1}}}\fb + \sqrt{1 - \alpha_{\tau_{s-1}}}\epsilonb}$ \;
            }
            $\mathcal{L}_{\textnormal{total}} \leftarrow\mathcal{L}_{\textnormal{direction}}(\hat\xb_0^{(i)},y_{\textnormal{tar}}; \xb_{0}^{(i)},y_{\textnormal{ref}})$\;
            $\mathcal{L}_{\textnormal{total}} \leftarrow\mathcal{L}_{\textnormal{total}}+\mathcal{L}_{\textnormal{id}}(\hat\xb_0^{(i)}, \xb_0^{(i)})$\;
        Take a gradient step on $\nabla_{\hat{\theta}} \mathcal{L}_{\textnormal{total}}.$
        }
        
    }
\end{algorithm}

 \begin{algorithm}[h!]
    \caption{GPU-efficient fine-tuning\label{algo:gpu_efficient_finetuning}}
    \DontPrintSemicolon
    \SetAlgoNoLine
    \SetAlgoVlined
    \SetKwProg{Fn}{Require}{:}{}
    \tcp{Step 2: Update the diffusion model}
    Clone the pretrained model $\epsilonb_{\hat{\theta}} \leftarrow \epsilonb_{{\theta}}$\;
    Define $\{ \tau_s\}^{S_{\textnormal{gen}}}_{s=1}$ s.t. $\tau_1=0,\tau_{  \scalebox{.8}{$\scriptscriptstyle S_{\textnormal{gen}} $}   }=t_0$. \;
    \For{$k = 1,2,\ldots, K$}{
        \For{$i = 1,2,\ldots, N$}{
            Clone the latent $\hat{\xb}^{(i)}_{t_0} \leftarrow \xb^{(i)}_{t_0}$. \;
            \For{$\ s = S_{\textnormal{gen}}, S_{\textnormal{gen}}-1,\ldots, 2$}{
            ${\epsilonb \leftarrow \epsilonb_{\hat{\theta}}(\hat{\xb}^{(i)}_{\tau_{s}}, \tau_{s}) ; \ \  \fb \leftarrow \fb_{\hat{\theta}}(\hat{\xb}^{(i)}_{\tau_{s}},\tau_{s})}$ \;
            ${\hat{\xb}^{(i)}_{\tau_{s-1}} \leftarrow \sqrt{\alpha_{\tau_{s-1}}}\fb + \sqrt{1 - \alpha_{\tau_{s-1}}}\epsilonb}$ \;
            $\mathcal{L}_{\textnormal{total}} \leftarrow\mathcal{L}_{\textnormal{direction}}(\fb,y_{\textnormal{tar}}; \xb_{0}^{(i)},y_{\textnormal{ref}})$\;
            $\mathcal{L}_{\textnormal{total}} \leftarrow\mathcal{L}_{\textnormal{total}}+\mathcal{L}_{\textnormal{id}}(\fb, \xb_0^{(i)})$\;
            Take a gradient step on $\nabla_{\hat{\theta}} \mathcal{L}_{\textnormal{total}}.$
            }
        }
        
    }
\end{algorithm}

\begin{figure}[!htb]
    \centering
    \vspace{-0.5em}
    \includegraphics[width=1.0\linewidth]{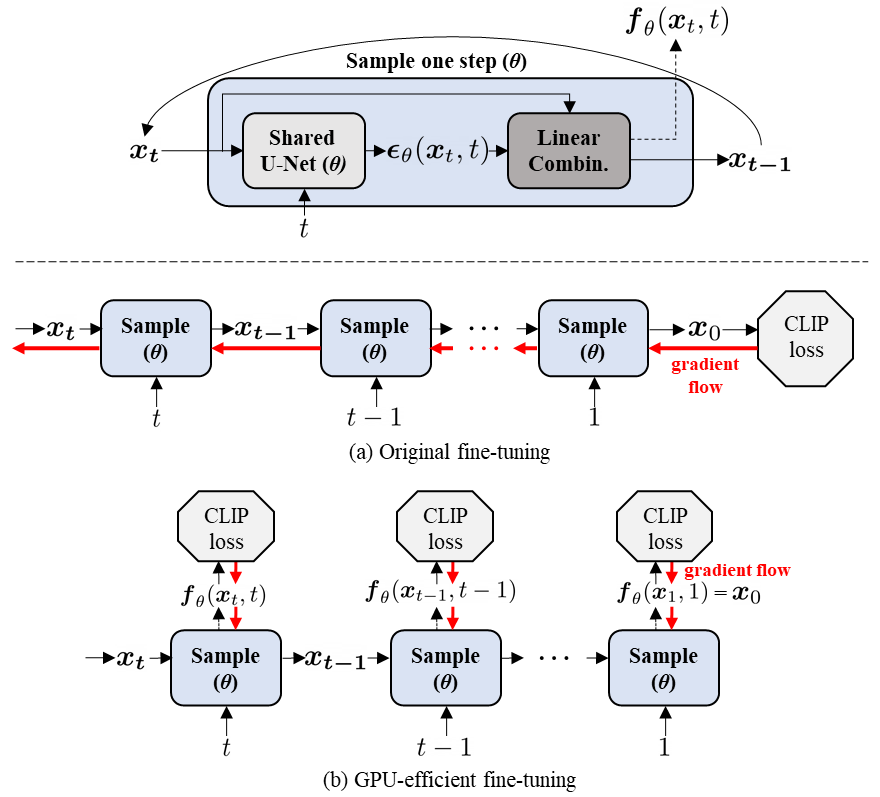}
    \vspace{-1em}
    \caption{Gradient flows during the fine-tuning and GPU-efficient fine-tuning. In GPU-efficient fine-tuning, loss calculation and a gradient step are proceeded at each time step $t$ with $\vf_\theta(\xb_{t}, t)$, the prediction of $\xb_0$ at $t$.}
    \vspace{-1em}
    \label{fig_s:efficient_fine_tuning}
\end{figure}

\paragraph{GPU-efficient fine-tuning.}
\label{app:gpu_efficient}
During the fine-tuning, the model is updated by back-propagating the gradient from the last step as illustrated in Fig.~\ref{fig_s:efficient_fine_tuning}(a) and Algorithm~\ref{algo:gpu_efficient_finetuning}. Although this method shows great manipulation performance, as the gradient pass the model $S_\text{gen}$ times, the GPU usage can be burdensome. Therefore, we additionally propose GPU-efficient fine-tuning method. {Here, as shown in in Fig.~\ref{fig_s:efficient_fine_tuning}(b), the back-propagation from the loss functions  is performed at each time step $t$. GPU-efficient fine-tuning can require {half of VRAM} usage compared to the original fine-tuning, but it requires {twice} as much time due to calculating loss and making steps at each time step. More details of running time can be found in Sec.~\ref{app:running_time}. We show the result of manipulating ImageNet~\cite{ILSVRC15} $512 \times 512$ images using GPU-efficient fine-tuning method in Fig.~\ref{fig_s:comparision_vqgan}, \ref{fig_s:imagenet_general}, \ref{fig_s:imagenet_tennis} and \ref{fig_s:imagenet_frog}.}

\paragraph{Image manipulation via fine-tuned model.}
Once the diffusion model $\epsilonb_{\hat{\theta}}$ is fine-tuned for the target control $y_\text{tar}$, the manipulation process of a input image $\xb_0$ is quite simple as in Algorithm~\ref{algo:manipulation}. Specifically, $\xb_0$ is inverted to $\xb_{t_0}$ through the forward DDIM process with the original pretrained model $\epsilonb_{{\theta}}$, followed by the reverse DDIM process with the fine-tuned model $\epsilonb_{\hat{\theta}}$ resulting $\hat{\xb}_0$. We use the same $t_0$ as used in the fine-tuning.

 \begin{algorithm}[!htb]
    \caption{DiffuisonCLIP manipulation\label{algo:manipulation}}
    \DontPrintSemicolon
    \SetAlgoNoLine
    \SetAlgoVlined
    \SetKwProg{Fn}{Require}{:}{}
    \SetKwProg{Fn}{Function}{:}{}
    \SetKwFunction{Manipulation}{Manipulation}
    \KwIn{$\xb_0$ (input image), $\epsilonb_{\hat{\theta}}$ (fine-tuned model), $\epsilonb_{{\theta}}$ (pretrained model), $t_0$ (return step), $S_{\textnormal{for}}$ (\# of inversion steps),  $S_{\textnormal{gen}}$(\# of generation steps)} 
    \BlankLine
    \Fn{\Manipulation{$\xb_0$, $\epsilonb_{\hat{\theta}}$, *}}
    {
    Define $\{ \tau_s\}^{S_{\textnormal{for}}}_{s=1}$ s.t. $\tau_1=0,\tau_{  \scalebox{.8}{$\scriptscriptstyle S_{\textnormal{for}} $}   }=t_0$. \;

    \For{$\ s = 1,2,\ldots, S_{\textnormal{for}}-1$}{
            ${\epsilonb \leftarrow \epsilonb_{{\theta}}(\xb_{\tau_{s}}, \tau_{s}) ; \ \  \fb \leftarrow \fb_\theta(\xb_{\tau_{s}},\tau_{s})}$ \;
            ${\xb_{\tau_{s+1}} \leftarrow \sqrt{\alpha_{\tau_{s+1}}}\fb + \sqrt{1 - \alpha_{\tau_{s+1}}}\epsilonb}$ \;
            }
    \BlankLine
    Define $\{ \tau_s\}^{S_{\textnormal{gen}}}_{s=1}$ s.t. $\tau_1=0,\tau_{  \scalebox{.8}{$\scriptscriptstyle S_{\textnormal{gen}} $}   }=t_0$ \;
    $\hat{\xb}_{t_0} \leftarrow \xb_{t_0}$\;
    \For{$\ s = S_{\textnormal{gen}}, S_{\textnormal{gen}}-1,\ldots, 2$}{
        ${\epsilonb \leftarrow \epsilonb_{\hat{\theta}}(\hat{\xb}_{\tau_{s}}, \tau_{s}) ; \ \  \fb \leftarrow \fb_{\hat{\theta}}(\hat{\xb}_{\tau_{s}},\tau_{s})}$ \;
        ${\hat{\xb}_{\tau_{s-1}} \leftarrow \sqrt{\alpha_{\tau_{s-1}}}\fb + \sqrt{1 - \alpha_{\tau_{s-1}}}\epsilonb}$ \; 
    }
    \KwRet{$\hat{\xb}_{0}$ \textnormal{(manipulated image)}}
    }
\end{algorithm}

\subsection{Image Translation between Unseen Domains}
By combining the method in SDEdit \cite{meng2021sdedit} and the manipulation with the fine-tuned model by DiffusionCLIP as detailed in Algorithm~\ref{algo:unseen2unseen}, we can even translate an image from an unseen domain into another unseen domain.
In the first step, the input image in the source unseen domain $\xb_0$ is first perturbed to $\xb'_{t_0}$ through the stochastic forward DDPM process \cite{ho2020denoising} until the return step $t_0$.
Next, the image in the pretrained domain $\xb'_{0}$ is sampled through the reverse DDIM process with the original pretrained model $\epsilonb_\theta$. These forward-generative processes are repeated for $K_\text{DDPM}$ times until the image $\xb'_{0}$ is close to the image in the pretrained domain. 

In the second step, $\xb'_{0}$ is manipulated into the image $\hat{\xb}_0$ in the CLIP-guided unseen domain with the fine-tuned model $\epsilonb_{\hat{\theta}}$ as in Algorithm~\ref{algo:manipulation}.

 \begin{algorithm}[!htb]
    \caption{Translation between unseen domains\label{algo:unseen2unseen}}
    \DontPrintSemicolon
    \SetAlgoNoLine
    \SetAlgoVlined
    \SetKwProg{Fn}{Require}{:}{}
    \SetKwProg{Fn}{Function}{:}{}
    \SetKwFunction{Manipulation}{Manipulation}
    \SetKwFunction{Unseen}{Unseen2Unseen}
    \KwIn{$\xb_0$ (image in an unseen domain or stroke), $\epsilonb_{\hat{\theta}}$ (fine-tuned model), $K_\textnormal{DDPM}$ (\# of iterations of Step 1), $\epsilonb_{{\theta}}$ (pretrained model), $t_0$ (return step), $S_{\textnormal{for}}$ (\# of inversion steps),  $S_{\textnormal{gen}}$(\# of generation steps)} 
    \KwOut{$\hat{\xb}_{0}$ \textnormal{(manipulated image)}}
    \tcp{Step 1: Source unseen $\rightarrow$ Pretrained}
    Define $\{ \tau_s\}^{S_{\textnormal{gen}}}_{s=1}$ s.t. $\tau_1=0,\tau_{  \scalebox{.8}{$\scriptscriptstyle S_{\textnormal{gen}} $}   }=t_0$. \;
    $\xb'_0 \leftarrow \xb_0 $ \;
    \For{$k = 1,2,\ldots, K_{\textnormal{DDPM}}$}{
        $\wb \sim \mathcal{N} (\mathbf{0,I})$ \;
        $\xb'_{t_0} \leftarrow \sqrt{\alpha_{t_0}}\xb'_0  + \sqrt{1 - \alpha_{t_0}}\bm{\wb}$  \;
        \For{$\ s = S_{\textnormal{gen}}, S_{\textnormal{gen}}-1,\ldots, 2$}{
        ${\epsilonb \leftarrow \epsilonb_{{\theta}}(\xb'_{\tau_{s}}, \tau_{s}) ; \ \  \fb \leftarrow \fb_{{\theta}}(\xb'_{\tau_{s}},\tau_{s})}$ \;
            ${\xb'_{\tau_{s-1}} \leftarrow \sqrt{\alpha_{\tau_{s-1}}}\fb + \sqrt{1 - \alpha_{\tau_{s-1}}}\epsilonb}$ \; 
    }
    }
    \tcp{Step 2: Pretrained $\rightarrow$ Target unseen}
    $\hat{\xb}_{0} \leftarrow$ \Manipulation{$\xb'_0$, $\epsilonb_{\hat{\theta}}$, *}
    
\end{algorithm}

\subsection{Noise Combination}
With the multiple diffusion models fine-tuned for the different controls $\{\epsilonb_{\hat{\theta}_i}\}^M_{i=1}$, we can change multiple attributes through only one sampling process. Specifically, we can flexibly mix several single attribute fine-tuned models with different combinations as described in Algorithm~\ref{algo:mlultiple}, without having to fine-tune new models with target texts that define multiple attributes. 

More specifically, we first invert an input image $\xb_0$ into $\xb_{t_0}$ via the forward DDIM process with the original pretrained diffusion model $\epsilonb_{{\theta}}$ as single attribute manipulation. Then, we use the multiple fine-tuned models during the reverse DDIM process. 
By applying different time dependent weight  $\gamma_i(t)$ satisfying  $\sum_{i=1}^M \gamma_i(t) = 1$ for each model, we can control the degree of change for multiple attributes.
 Of note, we can also apply this noise combination method for controlling the degree of change during single attribute manipulation. By mixing the noise from the original pretrained model $\epsilonb_{{\theta}}$ and the fine-tuned model $\epsilonb_{\hat{\theta}}$ concerning a single $\gamma$, we can perform interpolation between the original image and the manipulated image smoothly.

 \begin{algorithm}[h!]
    \caption{Multi-attribute transfer\label{algo:mlultiple}}
    \DontPrintSemicolon
    \SetAlgoNoLine
    \SetAlgoVlined
    \SetKwProg{Fn}{Require}{:}{}
    \SetKwProg{Fn}{Function}{:}{}
    \SetKwFunction{MultiTransfer}{MultiTransfer}
    \KwIn{$\xb_0$ (input image), $\{\epsilonb_{\hat{\theta}_i}\}^M_{i=1}$ (multiple fine-tuned models), $\epsilonb_{{\theta}}$ (pretrained model), $\{\gamma(t)_i\}^M_{i=1}$ (sequence of model weights), $t_0$ (return step), $S_{\textnormal{for}}$ (\# of inversion steps),  $S_{\textnormal{gen}}$(\# of generation steps)} 
    \KwOut{$\hat{\xb}_{0}$ \textnormal{(manipulated image)}}
    \BlankLine
    Define $\{ \tau_s\}^{S_{\textnormal{for}}}_{s=1}$ s.t. $\tau_1=0,\tau_{  \scalebox{.8}{$\scriptscriptstyle S_{\textnormal{for}} $}   }=t_0$. \;
    \For{$\ s = 1,2,\ldots, S_{\textnormal{for}}-1$}{
         ${\epsilonb \leftarrow \epsilonb_{{\theta}}(\xb_{\tau_{s}}, \tau_{s}) ; \ \  \fb \leftarrow \fb_\theta(\xb_{\tau_{s}},\tau_{s})}$ \;
         ${\xb_{\tau_{s+1}} \leftarrow \sqrt{\alpha_{\tau_{s+1}}}\fb + \sqrt{1 - \alpha_{\tau_{s+1}}}\epsilonb}$
    }
    
    Define $\{ \tau_s\}^{S_{\textnormal{gen}}}_{s=1}$ s.t. $\tau_1=0,\tau_{  \scalebox{.8}{$\scriptscriptstyle S_{\textnormal{gen}} $}   }=t_0$. \;

    $\hat{\xb}_{t_0} \leftarrow \xb_{t_0}$ \;
    \For{$\ s = S_{\textnormal{gen}}, S_{\textnormal{gen}}-1,\ldots, 2$}{
        $\epsilonb \leftarrow \sum_{i=1}^M \gamma_i({\tau_{s}}) \boldsymbol{\epsilon}_{\hat{\theta}_i}(\hat{\xb}_{\tau_{s}}, {\tau_{s}})$ \;
        $\fb \leftarrow \sum_{i=1}^M \gamma_i(\tau_{s}) \fb_{\hat{\theta}_i}(\hat{\xb}_{\tau_{s}}, {\tau_{s}})$ \;
        ${\hat{\xb}_{\tau_{s-1}} \leftarrow \sqrt{\alpha_{\tau_{s-1}}}\fb + \sqrt{1 - \alpha_{\tau_{s-1}}}\epsilonb}$ 
        
    }
\end{algorithm}

\section{Details on Network}
\label{app:network}
\begin{figure}[!htb]
    \centering
    \includegraphics[width=1.0\linewidth]{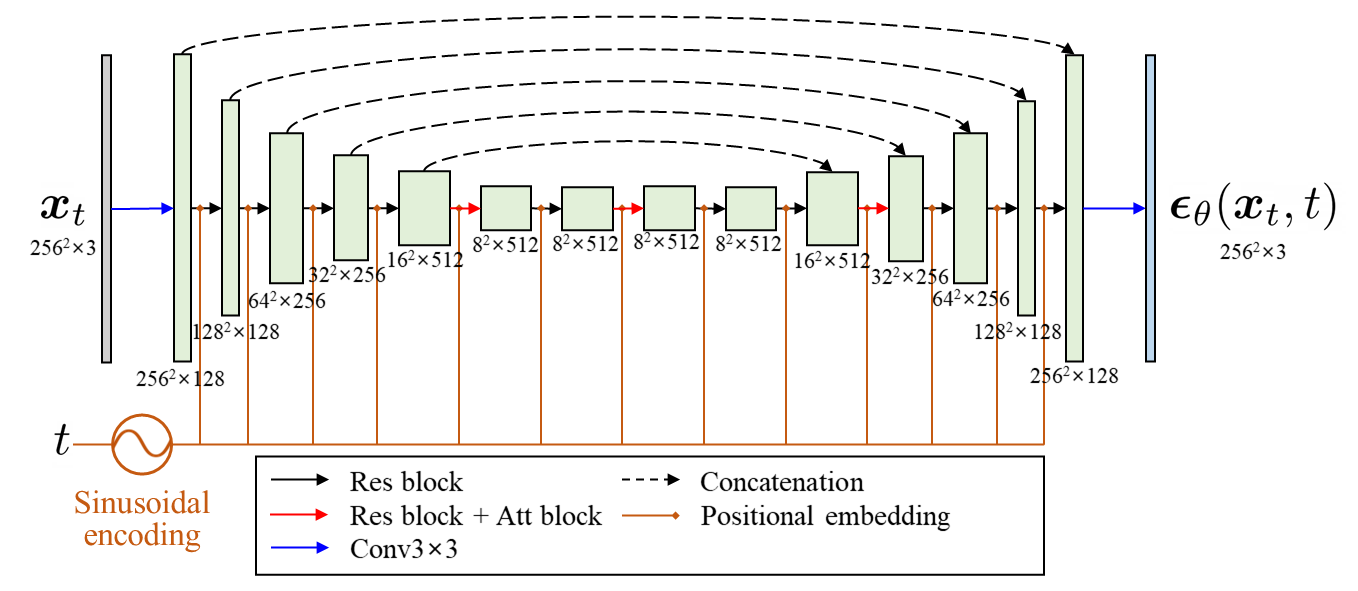}
    \caption{The shared U-Net architecture across $t$ of the diffusion model that generates $256 \times 256 $ images. The model receives $\xb_t$ and $t$ as inputs and outputs $\epsilonb_\theta(\xb_t,t)$.}
    \vspace{-1em}
    \label{fig_s:architecture}
\end{figure}

\begin{figure}[!htb]
    \centering
    \includegraphics[width=0.7\linewidth]{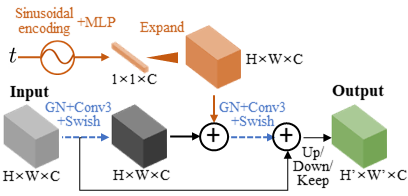}
    \vspace{-1em}
    \caption{Details of Res block.}
    \label{fig_s:resblock}
\end{figure}

Most of existing diffusion models receives $\xb_t$ and $t$ as inputs to the network $\epsilonb_\theta(\xb_t,t)$. We use the DDPM \cite{ho2020denoising} models pre-trained on $256 \times 256$ images in CelebA-HQ \cite{karras2017progressive}, LSUN-Bedroom and LSUN-Church \cite{yu15lsun} datasets. This model adopts the U-Net \cite{ronneberger2015u} architecture based on Wide-ResNet \cite{zagoruyko2016wide} shared across $t$ as represented in Fig.~\ref{fig_s:architecture}. In specific, the model is composed of the encoder part, middle part, decoder part, and time embedding part. In the encoder part, the $8 \times 8$ feature is generated from the $256 \times 256$ input image via 1 input convolution and 5 Res blocks. One Res block is composed of two convolutional blocks including Group normalization \cite{wu2018group} and Swish activation \cite{ramachandran2017searching} with the residual connection as in Fig.~\ref{fig_s:resblock}. At the $16 \times 16$ resolution, self-attention blocks are added to the Res block. The middle part consists of 3 Res blocks and the second block includes a self-attention block. In the decoder part, the output whose resolution is the same as the input is produced from the feature after the middle part through 5 Res blocks and 1 output convolution with skip connections from the features in the encoder part. In the time embedding part, the diffusion time $t$ is embedded into each Res blocks as represented in Fig.~\ref{fig_s:resblock} after the Transformer sinusoidal encoding as proposed in \cite{vaswani2017attention}. We use the models pretrained on Celeba-HQ, LSUN-Bedroom, and LSUN-Church models that are used in \cite{meng2021sdedit}.

For the manipulation of dog faces, we use the improved DDPM \cite{nichol2021improved} models pre-trained on AFHQ-Dog \cite{choi2020starganv2}. The architecture is almost same except that the model produces the extra outputs at the output convolution to predict the variance $\boldsymbol{\Sigma}_\theta(\xb_t,t)$ as well as the mean $\mub_\theta(\xb_t,t)$ which can be predicted from $\epsilonb_\theta(\xb_t,t)$. We use the models pretrained on AFHQ-Dog that is used in \cite{choi2021ilvr}. 

For the manipulation of $512 \times 512$ images from ImageNet dataset~\cite{ILSVRC15}, we use the improved DDPM \cite{nichol2021improved} pretrained model that is used in \cite{dhariwal2021diffusion}. Different from $256 \times 256$ resolution models, self-attention blocks are added to the Res block at the resolution of $8 \times 8, 16 \times 16$ and $32 \times 32$ resolution.


\section{Details and More Results of Comparison}
\label{app:comparision}

\subsection{Reconstruction}
{Here, we provide details on the quantitative comparison of reconstruction performance between our diffusion-based inversion and SOTA GAN inversion methods, which results are presented in Sec 4.1 and Tab. 1 of our main text. }

\paragraph{Baseline models.}
We use optimization approach \cite{abdal2019image2stylegan}, pixel2style2pixel (pSp) encoder \cite{richardson2021encoding}, Encoder for Editing (e4e) \cite{tov2021designing}, ReStyle encoder \cite{alaluf2021restyle} and HFGI encoder \cite{wang2021high} as our baseline models. pSp encoder adopts a Feature Pyramid Network and \cite{lin2017feature} inverts the image into $\mathcal{W}+$ space of StyleGAN.
In contrast, e4e converts the image to the latent in $\mathcal{W}$ space, which enables to explain the trade-offs between distortion and editing quality. Restyle encoder {is a residual-based encoder}, improving its performance using iterative refinement. HFGI encoder further improves the inversion performance leveraging the adaptive distortion alignment module and the distortion consultation module.

\paragraph{Comparison setting.}
We followed the experimental settings as described in \cite{wang2021high}. We invert the first 1,500 CelebA-HQ images. Then, we measure the quality of reconstruction from the inverted latent using MAE, LPIPS, SSIM metrics. All results except the result of our method are from the  \cite{wang2021high}. For our method, we set $(S_\text{for}, S_\text{gen})$ to (200, 40), which is our general setting.

\subsection{Human Evaluation}
\paragraph{Comparison setting.}
We conduct user study to evaluate real face image manipulation performance on CelebA-HQ~\cite{karras2017progressive} with our method, StyleCLIP global direction (GD) \cite{patashnik2021styleclip} and StyleGAN-NADA \cite{gal2021stylegan}. We get 6,000 votes from 50 people using a survey platform. {We use the first 20 images in CelebA-HQ testset as general cases and use another 20 images with novel views, hand pose, and fine details as hard cases.} For a fair comparison, we use 4 in-domain attributes (angry, makeup, beard, tanned) and 2 out-of-domain attributes (zombie, sketch), which are used in the studies of baselines. Here, we use official pretrained checkpoints and implementation for each approach.
We ask the respondents to rank the models by how well the image is manipulated, representing the property of the target attribute and preserving important semantics of the objects. 
\paragraph{Results used for evaluation.}
We provide manipulation results by our method, StyleCLIP-GD and StyleGAN-NADA, which are used for human evaluation, in Fig.~\ref{fig_s:human_eval_hard}, \ref{fig_s:human_eval_general}.

\subsection{Quantitative Evaluation}
\begin{figure}[!htb]
    \centering
    \vspace{-0.5em}
    \includegraphics[width=\linewidth]{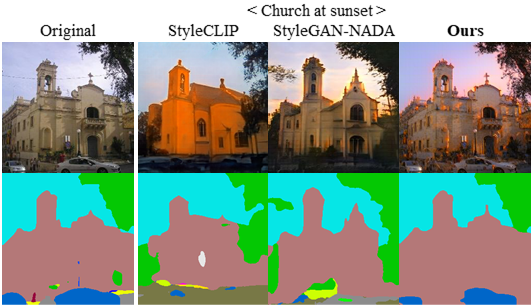}
    \vspace{-1em}
    \caption{Example of segmentation results from the manipulation results by different methods.}
    \vspace{-1em}
    \label{fig_s:segmentation}
\end{figure}

\paragraph{Quality metrics.}
{We use the following quality metrics for quantitative evaluation: Directional CLIP similarity ($\mathbf{\mathcal{S}_\text{dir}}$), segmentation-consistency (SC),
and face identity similarity (ID). Specifically, $\mathbf{\mathcal{S}_\text{dir}}$ is defined as follows:}
 \begin{eqnarray}
     \mathbf{\mathcal{S}_\text{dir}}\left(\xb_{\text{gen}},{y_{\text{tar}}}; \xb_{\text{ref}},{y_{\text{ref}}}\right): = \frac{\langle \Delta I, \Delta T\rangle}{\|\Delta I\|\|\Delta T\|},
     \end{eqnarray}
 where     \begin{eqnarray*}
&    \Delta T =  E_T{(y_{\text{tar}}}) - E_T({y_{\text{ref}}}),
    \Delta I =  E_I(\xb_{\text{gen}}) - E_I(\xb_{\text{ref}}).
 \end{eqnarray*}
{Here, $E_I$ and $E_T$ are CLIP’s image and text encoders, respectively, and
 $y_\text{tar}$, $\xb_\text{gen}$ are the text description of a target and the generated image, respectively. Also, ${y_{\text{ref}}}, \xb_{\text{ref}}$ denote the source domain text and image, respectively.}
 {Next, SC is a pixel accuracy when the segmentation result from $\xb_{\text{ref}}$ by the pretrained segmentation model is set as the label and the result from $\xb_\text{gen}$ is set as the prediction, as shown in Figure ~\ref{fig_s:segmentation}}.
 {Lastly, $\text{ID} := {L}_{\text{face}}(\xb_{\text{gen}}, \xb_{\text{ref}} )$ where $\Lc_{\text{face}}$ is the face identity loss in \cite{deng2019arcface}. 
 
 Our goal is to achieve the better score in terms of $\mathbf{\mathcal{S}_\text{dir}}$, SC, and ID to demonstrate high attribute-correspondence ($\mathcal{S}_\text{dir}$) as well as well-preservation of identities without unintended changes (SC, ID).}

\paragraph{Comparison setting.}
{To compute $\mathbf{\mathcal{S}_\text{dir}}$, we use a pretrained CLIP~\cite{radford2021learning}. To calculate SC, we use pretrained face parsing network~\cite{yu2018bisenet} and semantic segmentation networks~\cite{zhou2017scene, zhou2018semantic}. To compute ID, we use a pretrained face recognition~\cite{deng2019arcface} model.
Then, we performed comparison with StyleCLIP~\cite{patashnik2021styleclip} and StyleGAN-NADA~\cite{gal2021stylegan}.
We use 1,000 test images from CelebA-HQ~\cite{karras2017progressive} and LSUN-Church~\cite{yu15lsun}, respectively.
We use the manipulation results for three attributes in CelebA-HQ (makeup, tanned, gray hair) and  LSUN-Church (golden, red brick, sunset). These attributes
are required to confirm that the manipulation results correspond to the target text without the changes of identities and shapes of the source objects.}


\subsection{Comparison of Church Image Manipulation}
We additionally provide the manipulation of $256 \times 256$ church images from LSUN-Church \cite{yu15lsun} with StyleCLIP latent optimization (LO) \cite{patashnik2021styleclip} and StyleGAN-NADA \cite{gal2021stylegan} in Fig.~\ref{fig_s:comparision_church}.

\subsection{Diffusion-based Manipulations}
\begin{figure}[!htb]
    \centering
    \vspace{-0.5em}
    \includegraphics[width=\linewidth]{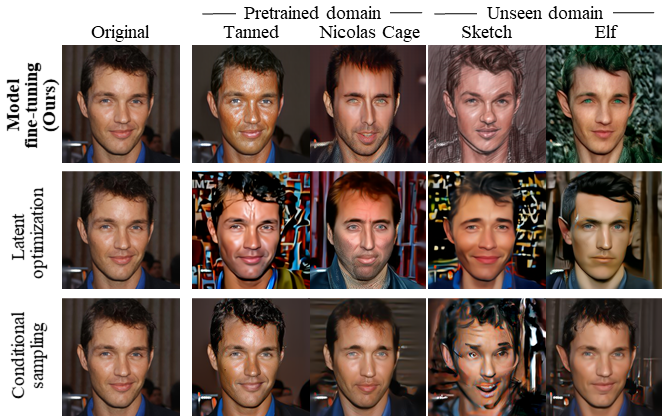}
    \vspace{-1em}
    \caption{Comparison between diffusion-based manipulation methods.}
    \vspace{-1em}
    \label{fig_s:comparision_diffusion}
\end{figure}

We compare our model fine-tuning method with latent optimization and conditional sampling method \cite{dhariwal2021diffusion} guided by CLIP loss.

For the latent optimization of the diffusion models, we use the same objective (Eq. (10) in the main manuscript) as the model fine-tuning. However, we optimize the inverted latent $\hat{\xb}_{t_0}$ instead of the model $\epsilonb_{\hat{\theta}}$. For conditional sampling, the sampling process is guided by the gradient of CLIP loss with respect to the latent as a classifier guides the process in \cite{dhariwal2021diffusion}. This method requires a noisy classifier that can classify the image with noise, but the noisy CLIP model is not publicly available and its training will be too expensive. To mitigate this issue, we use the method proposed by \cite{cgd2021clay}. Instead of using noisy CLIP, they use the gradient from the normal CLIP loss with the predicted $\xb_0$ given $\xb_t$, which we denoted as $\fb_\theta(\xb_t, t)$ in Eq. (\ref{eq:f}) at every step.

In Fig. \ref{fig_s:comparision_diffusion}, we display a series of the real image manipulation given the text prompt by our model fine-tuning method, latent optimization {and conditional sampling}. We can see that the manipulation results via latent optimization and conditional sampling methods failed to manipulate the images to the unseen domain. The reason is that the manipulation using latent optimization {and conditional sampling} is restricted by the learned distribution of the pretrained model. On the other hand, the proposed model fine-tuning method shows superior manipulation performance. 

\subsection{Other GAN Baselines}
\paragraph{Comparison with VQGAN-CLIP.}

\begin{figure}[!htb]
    \centering
    \vspace{-0.5em}
    \includegraphics[width=1.0\linewidth]{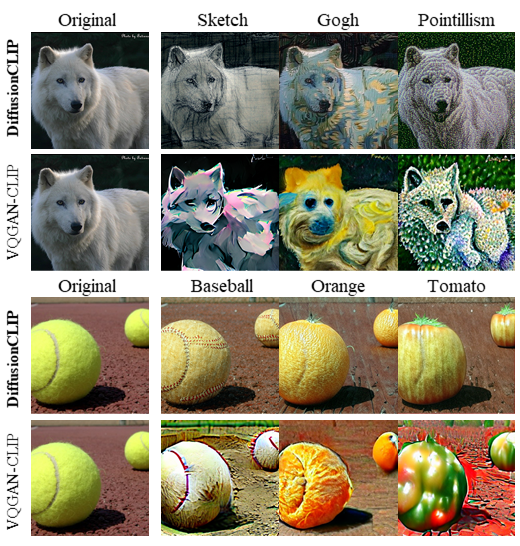}
    \vspace{-1em}
    \caption{Comparison  with VQGAN-CLIP~\cite{esser2020taming, radford2021learning} using $512 \times 512$ images from ImageNet~\cite{ILSVRC15} }
    \vspace{-1em}
    \label{fig_s:comparision_vqgan}
\end{figure}
VQGAN-CLIP~\cite{esser2020taming, radford2021learning} recently show the impressive results of CLIP-guided conditional generation of artistic images. It also provides the style transfer, which optimizes the latent from the input image guided by CLIP loss. We compare DiffusionCLIP with VQGAN-CLIP for the manipulation of $512 \times 512$ images from ImageNet~\cite{ILSVRC15}. We follow official implementation for VQGAN-CLIP. For our method, we utilize GPU-efficient fine-tuning method with the diffusion model pretrained on $512 \times 512$ ImageNet which is used in \cite{dhariwal2021diffusion}. We set $(S_\text{for}, S_\text{gen}) = (40, 12)$.
In the first two rows of Fig.~\ref{fig_s:comparision_vqgan}, our method successfully translates the image into target style, preserving the identity of the object. However, the manipulation results by VQGAN-CLIP do not show representative properties of target styles. In the bottom two rows of Fig.~\ref{fig_s:comparision_vqgan}, our method shows excellent semantic manipulation results preserving the details of backgrounds, while the results from VQGAN-CLIP show severe unintended changes.

\paragraph{Other GAN inversion-based manipulation.}
\begin{figure}[!htb]
    \centering
    \vspace{-0.5em}
    \includegraphics[width=\linewidth]{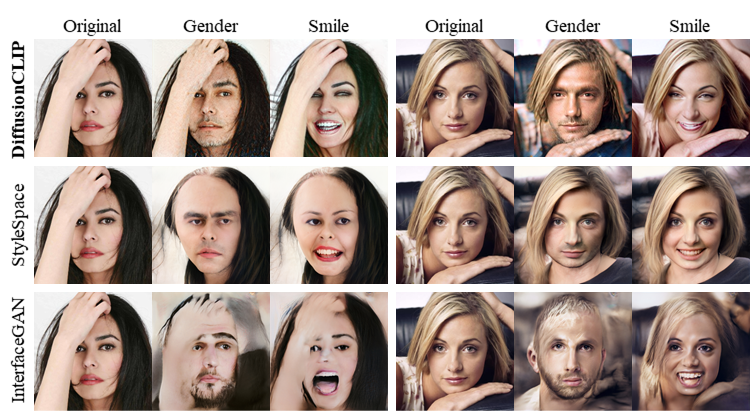}
    \vspace{-1em}
    \caption{Comparision with other GAN inversion-based manipulation: StyleSpace \cite{wu2021stylespace} and InterfaceGAN \cite{shen2020interfacegan}.}
    \label{fig_s:comparision_other_gan}
\end{figure}

We also compare our method with non text-driven manipulation methods based on GAN inversion: StyleSpace \cite{wu2021stylespace} and InterfaceGAN \cite{shen2020interfacegan}. StyleSpace manipulates the latent inverted by e4e \cite{tov2021designing} in StyleGAN2 \cite{karras2020analyzing} $\mathcal{W+}$ space. InterfaceGAN manipulates the latent inverted by IDInvert \cite{zhu2020domain} in StyleGAN~\cite{karras2019style} $\mathcal{W+}$ space. As shown in Fig.~\ref{fig_s:comparision_other_gan}, StyleSpace and InterfaceGAN fail to manipulate the images with hand poses, suggesting practical limitations. However, our method successfully manipulates the images without artifacts.

\section{Additional Results}
\label{app:additional_results}
\paragraph{Manipulation of $\bm{512 \times 512}$ images from ImageNet.}
Here, we provide the results of the manipulation of $512 \times 512$ images from ImageNet~\cite{ILSVRC15}. We leverage GPU-efficient fine-tuning with the diffusion model pretrained on $512 \times 512$ ImageNet which is used in \cite{dhariwal2021diffusion}. We set $(S_\text{for}, S_\text{gen}) = (40, 12)$. We set $(S_\text{for}, S_\text{gen}) = (40, 12)$ and other hyperparameters are equally applied as manipulation of $256 \times 256$ images. We first show the style transfer results of general images in Fig.~\ref{fig_s:imagenet_general}. We show text-driven semantic manipulation results from  tennis ball into other objects in Fig.~\ref{fig_s:imagenet_tennis}. Finally, we show the manipulation of frog images in Fig.~\ref{fig_s:imagenet_frog}.

\paragraph{Image translation between unseen domains.}
In Fig.~\ref{fig_s:additional_unseen2unseen} we display additional results of image translation between unseen domains, where animation images, portrait art, and strokes are translated into Pixar, paintings by Gogh and Neanderthal men. Note that
we do not require any curated dataset for both source and target domain.

\paragraph{Failure cases.}
\begin{figure}[!htb]
    \centering
    \vspace{-0.5em}
    \includegraphics[width=\linewidth]{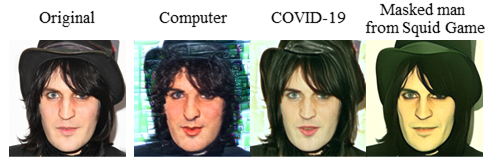}
    \vspace{-1em}
    \caption{Failure cases.}
    \vspace{-1em}
    \label{fig_s:failure_cases}
\end{figure}

Due to the dependency on the performance of CLIP encoder, DiffusionCLIP sometimes fails to manipulate images as shown in Fig.~\ref{fig_s:failure_cases}. For example, it is difficult to manipulate human face images into objects such as computers, chairs, pencils. Also, manipulation to target controls that happen or become famous recently may fail because their representations are not reflected inside CLIP encoders.

\section{Hyperparameter and Ablation Study}
\label{app:hyperparameter_ablation}
\subsection{Dependency on $S_\text{for}, S_\text{gen}$ and $t_{0}$}
 In Table~\ref{tab:recon_wrt_hyperparam}, the reconstruction from the latents through the inversion process on face images are evaluated using MAE, LPIPS and SSIM. As $S_\text{for}$ and $S_\text{gen}$ increase, the reconstruction quality increases. However, in case that $S_\text{for} < S_\text{gen}$, the quality stays in the similar degree or even decreases, causing the artifacts as the cases of $(S_\text{for}, S_\text{gen})=(6,40)$ and $(S_\text{for}, S_\text{gen})=(200,6)$ in Fig. 10 in the main  manuscript. When $(S_\text{for}, S_\text{gen})$ is fixed, as the return step $t_0$ increases, the quality decreased because the intervals between the steps become larger.
 
 \begin{table*}[!htb]
    \caption{Quantitative analysis on reconstruction quality with respect to $S_\text{for}, S_\text{gen}$ and $t_{0}$.}\label{tab:recon_wrt_hyperparam}
    \centering
    \begin{adjustbox}{width=0.7\linewidth}
    
        \begin{tabular}{cccccc}
            \toprule
            $t_0$ & $S_\text{for}$ & $S_\text{gen}$ & MAE ↓ & LPIPS ↓ & SSIM ↑ \\
            \midrule
            \multirow{9}{*}{300} & \multirow{3}{*}{6}   & 6   & 0.047 & 0.185 & 0.732 \\
                                                     &  & 40  & 0.061 & 0.221 & 0.704 \\
                                                     &  & 200 & 0.063 & 0.224 & 0.694 \\
                                                     \cmidrule{2-6}
                                 & \multirow{3}{*}{40}  & 6   & 0.027 & 0.110 & 0.863 \\
                                                     &  & 40  & 0.023 & 0.091 & 0.891 \\
                                                     &  & 200 & 0.023 & 0.086 & 0.895 \\
                                                     \cmidrule{2-6}
                                 & \multirow{3}{*}{200} & 6   & 0.024 & 0.095 & 0.885 \\
                                                     &  & 40  & 0.020 & 0.073 & 0.914 \\
                                                     &  & 200 & 0.019 & 0.065 & 0.923 \\
                                                     \midrule
            \multirow{9}{*}{400} & \multirow{3}{*}{6}   & 6   & 0.055 & 0.208 & 0.673 \\
                                                     &  & 40  & 0.073 & 0.255 & 0.655 \\
                                                     &  & 200 & 0.077 & 0.260 & 0.643 \\
                                                     \cmidrule{2-6}
                                & \multirow{3}{*}{40}   & 6   & 0.031 & 0.128 & 0.827 \\
                                                     &  & 40  & 0.025 & 0.100 & 0.880 \\
                                                     &  & 200 & 0.024 & 0.093 & 0.885 \\
                                                     \cmidrule{2-6}
                                & \multirow{3}{*}{200}  & 6   & 0.028 & 0.108 & 0.862 \\
                                                     &  & 40  & 0.024 & 0.076 & 0.910 \\
                                                     &  & 200 & 0.020 & 0.068 & 0.919 \\
             \bottomrule
        \end{tabular}
    
    \quad

        \begin{tabular}{cccccc}
            \toprule
            $t_0$ & $S_\text{for}$ & $S_\text{gen}$ & MAE ↓ & LPIPS ↓ & SSIM ↑ \\
            \midrule
            \multirow{9}{*}{500} & \multirow{3}{*}{6}   & 6   & 0.065 & 0.237 & 0.602 \\
                                                     &  & 40  & 0.085 & 0.286 & 0.615 \\
                                                     &  & 200 & 0.090 & 0.292 & 0.602 \\
                                                     \cmidrule{2-6}
                                 & \multirow{3}{*}{40}  & 6   & 0.037 & 0.148 & 0.779 \\
                                                     &  & 40  & 0.027 & 0.109 & 0.868 \\
                                                     &  & 200 & 0.026 & 0.101 & 0.874 \\
                                                     \cmidrule{2-6}
                                 & \multirow{3}{*}{200} & 6   & 0.032 & 0.126 & 0.827 \\
                                                     &  & 40  & 0.022 & 0.082 & 0.901 \\
                                                     &  & 200 & 0.021 & 0.073 & 0.912 \\
                                                     \midrule
            \multirow{9}{*}{600} & \multirow{3}{*}{6}   & 6   & 0.084 & 0.283 & 0.501 \\
                                                     &  & 40  & 0.101 & 0.325 & 0.564 \\
                                                     &  & 200 & 0.106 & 0.330 & 0.552 \\
                                                     \cmidrule{2-6}
                                & \multirow{3}{*}{40}   & 6   & 0.047 & 0.175 & 0.706 \\
                                                     &  & 40  & 0.029 & 0.120 & 0.852 \\
                                                     &  & 200 & 0.028 & 0.108 & 0.862 \\
                                                     \cmidrule{2-6}
                                & \multirow{3}{*}{200}  & 6   & 0.041 & 0.147 & 0.778 \\
                                                     &  & 40  & 0.024 & 0.087 & 0.893 \\
                                                     &  & 200 & 0.022 & 0.076 & 0.907 \\
             \bottomrule
        \end{tabular}
    \end{adjustbox}
\end{table*}

\label{app:detailed_ablation}
\subsection{Identity Loss}

\begin{figure}[!htb]
    \centering
    \vspace{-0.5em}
    \includegraphics[width=\linewidth]{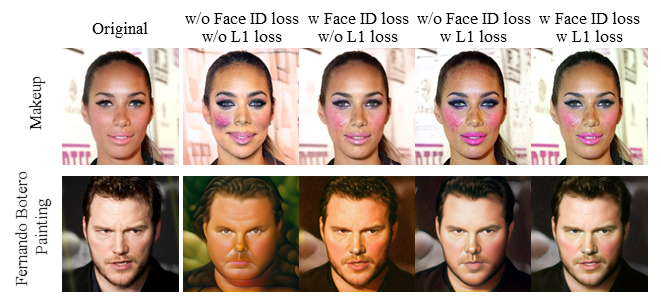}
    \vspace{-1em}
    \caption{Ablation study of identity loss.}
    \vspace{-1em}
    \label{fig_s:ablation_identitiy}
\end{figure}

Here, we analyze the importance of identity loss. We use $\ell_1$ loss as the identity loss, and in the case of human face image manipulation, the face identity loss in \cite{deng2019arcface} is used. Whether to use these identity losses is determined by the target control. We show the examples in Fig.~\ref{fig_s:ablation_identitiy}. If preserving the identity of the human face is important for the target control such as `Makeup', it is recommended to use face identity loss as we can see in the first row in Fig.~\ref{fig_s:ablation_identitiy}. $\ell_1$ can help further preserve the background details.  If the target control doesn't require the exact identity preserving as artistic transfer as the second rows of Fig.~\ref{fig_s:ablation_identitiy}, the identity loss can hinder the change. The examples of usage of hyperparameters depending on the target text prompts are represented in Table~\ref{tab:hyperparameter}.

\subsection{Dependency on Fine-tuning Epochs $K$}

\begin{figure}[!htb]
    \centering
    \includegraphics[width=\linewidth]{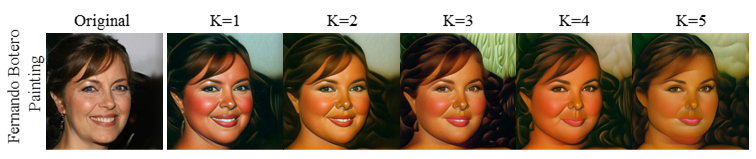}
    \vspace{-1em}
    \caption{Changes according to the fine-tuning epochs.}
    \label{fig_s:changes_wrt_epochs}
\end{figure}

To fine-tune diffusion models, we use Adam optimizer with an initial learning rate of 4e-6 which is increased linearly by 1.2 per 50 iterations. Hence, as we can see in the example of changes are represented in Fig.~\ref{fig_s:changes_wrt_epochs}, the images generated from the fine-tuned models change closer to the target control as the epoch $K$ increases.

\subsection{Dependency on the Number of Precomputed Images $N$}

\begin{figure}[!htb]
    \centering
    \includegraphics[width=\linewidth]{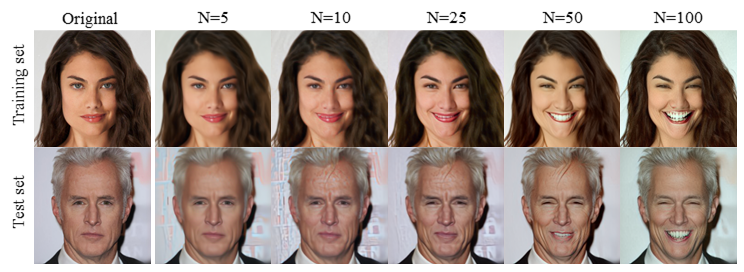}
    \vspace{-1em}
    \caption{Dependency on the number of precomputed images $N$}
    \label{fig_s:num_precomputed_images}
\end{figure}

As we mentioned before, if several latents have been precomputed, we can further reduce the time for fine-tuning by recycling the latent to synthesize other attributes. In this case, the number of precomputed images $N$ is a hyperparameter to be controlled. We test the cases with different $N$. We fine-tune the models with $N=5, 10, 25, 50, 100$, fixing the learning rates to 4e-6 and the number of iterations to 100. We found that as increasing the $N$, the image can be manipulated more as shown as Fig. \ref{fig_s:num_precomputed_images}.

\subsection{Stochastic Manipulation}

\begin{figure}[!htb]
    \centering
    \vspace{-1em}
    \includegraphics[width=\linewidth]{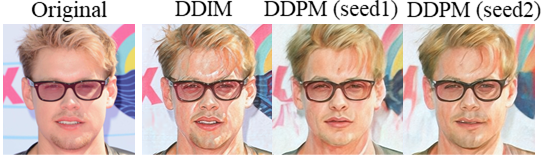}
    \vspace{-1em}
    \caption{Effect of stochastic manipulation with random seeds.}
    \vspace{-1em}
    \label{fig_s:stochastic_manipulation}
\end{figure}

{We analyzed how the results change when stochastic DDPM sampling is used rather than deterministic DDIM sampling. As shown in Figure~\ref{fig_s:stochastic_manipulation}, the images can be modified in many ways, which can be useful for artistic transfer.}

\subsection{Hyperparameters according to Target Text $y_\text{tar}$}

We provide examples of hyperparameter settings according to $y_\text{tar}$ in Table \ref{tab:hyperparameter}. {Our method has a similar number of hyperparameters as other text-driven methods such as StyleCLIP~\cite{patashnik2021styleclip} and StyleGAN-NADA~\cite{gal2021stylegan}.
In our method, the actual hyperparameters for different controls are just $t_0$, $\lambda_\text{L1}$, $\lambda_\text{ID}$. These can be chosen simply based on insight as to whether the target requires severe shape changes.} The target controls demanding severe changes of shape or color such as change of species or artistic style transfer require high $t_0$ without no identity losses, while the target controls were preserving the identity of the object is important to require low $t_0$ and the use of identity losses.

\begin{table}[h!]
\centering
\caption{Examples of hyperparameter settings according to $y_\text{tar}$.}\label{tab:hyperparameter}%
\begin{adjustbox}{width=\linewidth}
\begin{tabular}{cllcccc}
\toprule
{Type} & {$y_\text{tar}$} & {$y_\text{ref}$} & {$t_0$} & {$\lambda_\text{L1}$} & $\lambda_\text{ID}$ \\ 
 \midrule
  \multirow{20}{*}{Human face} & \textit{Tanned face} & \textit{face} & 300 & 0.3 & 0.3 \\
 & \textit{Face with makeup} & \textit{Face} & 300 & 0.3 & 0.3  \\
 & \textit{Face without makeup} & \textit{Face} & 300 & 0.3 & 0.3  \\
 & \textit{Angry face} & \textit{face} & 500 & 0.3 & 0.3  \\
 & \textit{Person with beards} & \textit{Person} & 400 & 0.3 & 0.3  \\
 & \textit{Person with curly hair} & \textit{Person} & 400 & 0.3 & 0.3 \\
 & \textit{Person with red hair} & \textit{Person} & 500 & 0.3 & 0.3  \\
 & \textit{Person with grey hair} & \textit{Person} & 500 & 0.3 & 0.3  \\
 & \textit{Old person} & \textit{person} & 400 & 0.3 & 0.3 \\
 & \textit{Mark Zuckerberg} & \textit{Person} & 600 & 0.3 & 0  \\
 & \textit{Painting by Gogh} & \textit{photo} & 600 & 0 & 0 \\
 & \textit{Painting in Modigliani style} & \textit{Photo} & 600 & 0 & 0 \\
 & \textit{Sketch} & \textit{Photo} & 600 & 0.3 & 0  \\
 & \textit{3D render in the style of Pixar} & \textit{Photo} & 600 & 0.3 & 0  \\
 & \textit{Portrait by Firda Kahlo} & \textit{Photo} & 600 & 0.3 & 0  \\
 & \textit{Super Saiyan} & \textit{Human} & 600 & 0 & 0  \\
 & \textit{Tolken elf} & \textit{Human} & 600 & 0 & 0 \\
 & \textit{Zombie} & \textit{Human} & 600 & 0 & 0 \\
 & \textit{The Jocker} & \textit{Human} & 600 & 0 & 0  \\
 & \textit{Neanderthal} & \textit{Human} & 600 & 0 & 0  \\
 \midrule
\multirow{10}{*}{Dog face} & \textit{Smiling Dog} & \textit{Dog} & 600 & 0.3 & -\\
 & \textit{Yorkshire Terrier} & \textit{Dog} & 600 & 0 & -  \\
 & \textit{Hamster} & \textit{Dog} & 600 & 0 & -  \\
 & \textit{Bear} & \textit{Dog} & 500 & 0 & - \\
 & \textit{Wolf} & \textit{Dog} & 500 & 0 & -  \\
 & \textit{Fox} & \textit{Dog} & 500 & 0 & -  \\
 & \textit{Nicolas Cage} & \textit{Dog} & 600 & 0 & - \\
 & \textit{Zombie} & \textit{Dog} & 600 & 0.3 & -  \\
 & \textit{Venom} & \textit{Dog} & 600 & 0.3 & -  \\
 & \textit{Painting by Gogh} & \textit{Photo} & 500 & 0.3 & - \\
 \midrule
\multirow{6}{*}{Church} & \textit{Red brick wall church} & \textit{church} & 300 & 0.3 & -  \\
 & \textit{Golden church} & \textit{church} & 400 & 0.3 & -  \\
 & \textit{Snow covered church} & \textit{church} & 500 & 0.3 & -  \\
 & \textit{Wooden house} & \textit{church} & 500 & 0.3 & - \\
 & \textit{Ancient traditional Asian tower} & \textit{church} & 500 & 0.3 & -  \\
 & \textit{Departmetn store} & \textit{church} & 500 & 0.3 & - \\
 \midrule
\multirow{6}{*}{Bedroom} & \textit{Blue tone bedroom} & \textit{bedroom} & 500 & 0.3 & -  \\
 & \textit{Green tone bedroom} & \textit{bedroom} & 500 & 0.3 & -  \\
 & \textit{Wooden bedroom} & \textit{bedroom} & 500 & 0.3 & - \\
 & \textit{Golden bedroom} & \textit{bedroom} &	400	& 0.3 & - \\
 & \textit{Palace bedroom} & \textit{bedroom} & 500 & 0.3 & - \\
 & \textit{Princess bedroom} & \textit{bedroom} & 500 & 0.3 & -  \\
 & \textit{Watercolor art with thick brushstrokes} & \textit{Photo} & 600 & 0.3 & - \\
  \bottomrule

\end{tabular}
\end{adjustbox}
\end{table}

\section{Running Time and Resources}
\label{app:running_time}
Here, we provide the details on the running time of training and inference for each procedure using NVIDIA Quadro RTX 6000 in the case of manipulating $256 \times 256$ size images.

\paragraph{DiffuisonCLIP fine-tuning.}
As illustrated in Sec~\ref{app:method_fine}, DiffusionCLIP fine-tuning process can be split into the latent precomputing procedure and the model updating procedure. The latent precomputing procedure is carried out just once for the same pre-trained diffusion. When we use $S_\text{for}$ of 40 as normal, the inversion for each image takes 1.644 seconds (all the reported times are the average times of 30 iterations). So, when we precompute the latents from the 50 images, it finished at about 82.2 seconds. For the model updating process, one update step including the generative process, loss calculation, and taking a gradient step takes 0.826 seconds when the batch size is 1 and $S_\text{gen}$ is 6. So, 1 epoch with 50 precomputed image-latent pairs takes 41.3 seconds. The total epochs $K$ are range from 1 to 10 depending on types of the target text $y_\text{tar}$, so the total time for the model updating takes from 41.3 seconds to 7 minutes.

{When using GPU-efficient model updating, loss calculation and taking a gradient step takes 1.662 seconds which is almost twice as the original fine-tuning. Therefore, total fine-tuning time will be increased as twice.}

The latent precomputing procedure requires about 6GB. The original model and GPU-efficient model updating require 23GB and 12GB of VRAM, respectively.


\paragraph{Manipulation of images from pretrained domain.}
With the quick manipulation $(S_{\text{for}}, S_\text{gen})=(40, 6)$, it takes 1.644 seconds and 0.314 seconds for the inversion process and the generative process, respectively, resulting in 1.958 seconds total. The quick manipulation still produces great results that can be well used for image manipulation in practice. 
When we set $(S_\text{for}, S_\text{gen})$ to (200, 40), it takes 8.448 seconds and 1.684 seconds for the inversion process and the generative process respectively, leading to 10.132 seconds in total. This application and the following applications all require at least 6GB of VRAM.

\paragraph{Image translation between unseen domains.}
Image translation between unseen domains and stroke-conditioned unseen domain generation requires $K_\text{DDPM}$  forward DDPM and reverse DDIM process added to one forward and reverse DDIM process. Thanks to the possibility of the sampling $\xb'_t$ in closed form, the time for forward DDPM and reverse DDIM process can be reduced into the time for the reverse DDIM process 0.314 seconds when $S_\text{gen}=6$. $K_\text{forward}$ is set to 1-10, so $K_\text{DDPM}$ forward DDPM and reverse DDIM process takes 0.314-3.14 seconds. When time for one forward and reverse DDIM process is added, the whole process takes  2.272-5.098 seconds with $(S_\text{for}, S_\text{gen}) = (40,6)$ and 10.446-13.272 seconds with $(S_\text{for}, S_\text{gen}) = (200,40)$.

\paragraph{Multi-attribute transfer.}
We can change multiple attributes through only one generative process. It takes 2.602 seconds when $(S_\text{for}, S_\text{gen}) = (40,6)$ and  14.744 seconds when $(S_\text{for}, S_\text{gen} = (200,40)$. 

\paragraph{Trade-off between the inference time and preparation time.} 
Latent optimization-based manipulation methods \cite{patashnik2021styleclip} do not require the preparation time for the manipulation. However, they require an optimization process per image. In contrast, our fine-tuning methods, latent mapper in StyleCLIP \cite{patashnik2021styleclip} and StyleGAN-NADA \cite{gal2021stylegan} require the set-up for manipulation, which is training the model. However, once the model is fine-tuned, we can apply the model to all images from the same pretrained domain. In terms of training time, our method takes 1-7 minutes, which is faster than the latent mapper of StyleCLIP (10-12hours) and similar to StyleGAN-NADA (a few minutes).

\paragraph{Increasing image size.}
{We found that as the image size is increased from $256 \times 256$ to $512 \times 512$, the running time for each procedure increased as 4 times, and GPU usage increased as twice.}

\section{Societal Impacts}
\label{app:societal_effect}
DiffusionCLIP enables high-quality manipulation of images for people using simple text prompts without professional artistic skills. However, this manipulation can be used maliciously to confuse people with realistic manipulated results. Therefore, we advise users to make use of our method properly.  We also advise you to make use of our method carefully for proper purposes. 

In this work, we use two types of pretrained models, CLIP~\cite{radford2021learning} and the diffusion models, to manipulate images without additional manual efforts for new target controls. Image encoder and text encoder of CLIP are trained on 400 million image-text pairs gathered from publicly available sources on the internet to learn visual concepts from natural language supervision. However, although the size of the training dataset is huge, it is not enough for the models to learn general balanced knowledge. As the authors in \cite{radford2021learning} acknowledged the potential issues from model biases, manipulation using CLIP can introduce biased results. Diffusion models trained on CelebA-HQ\cite{karras2017progressive}, AFHQ-dog~\cite{choi2020starganv2}, LSUN-Bedroom, LSUN-Church~\cite{yu15lsun} and ImageNet~\cite{ILSVRC15} used in our models can generate biased results during iterations. Especially, the generative models trained on the CelebA-HQ dataset that is composed of face images of celebrities are founded to produce face images of attractive people who are mostly 20-40 years old \cite{esser2020note}. We hope that more research is conducted in direction of generative models and representation learning that resolve the bias issues.



\clearpage

\begin{figure*}[!htb]
    \centering
    \vspace{-2em}
    \includegraphics[width=0.85\linewidth]{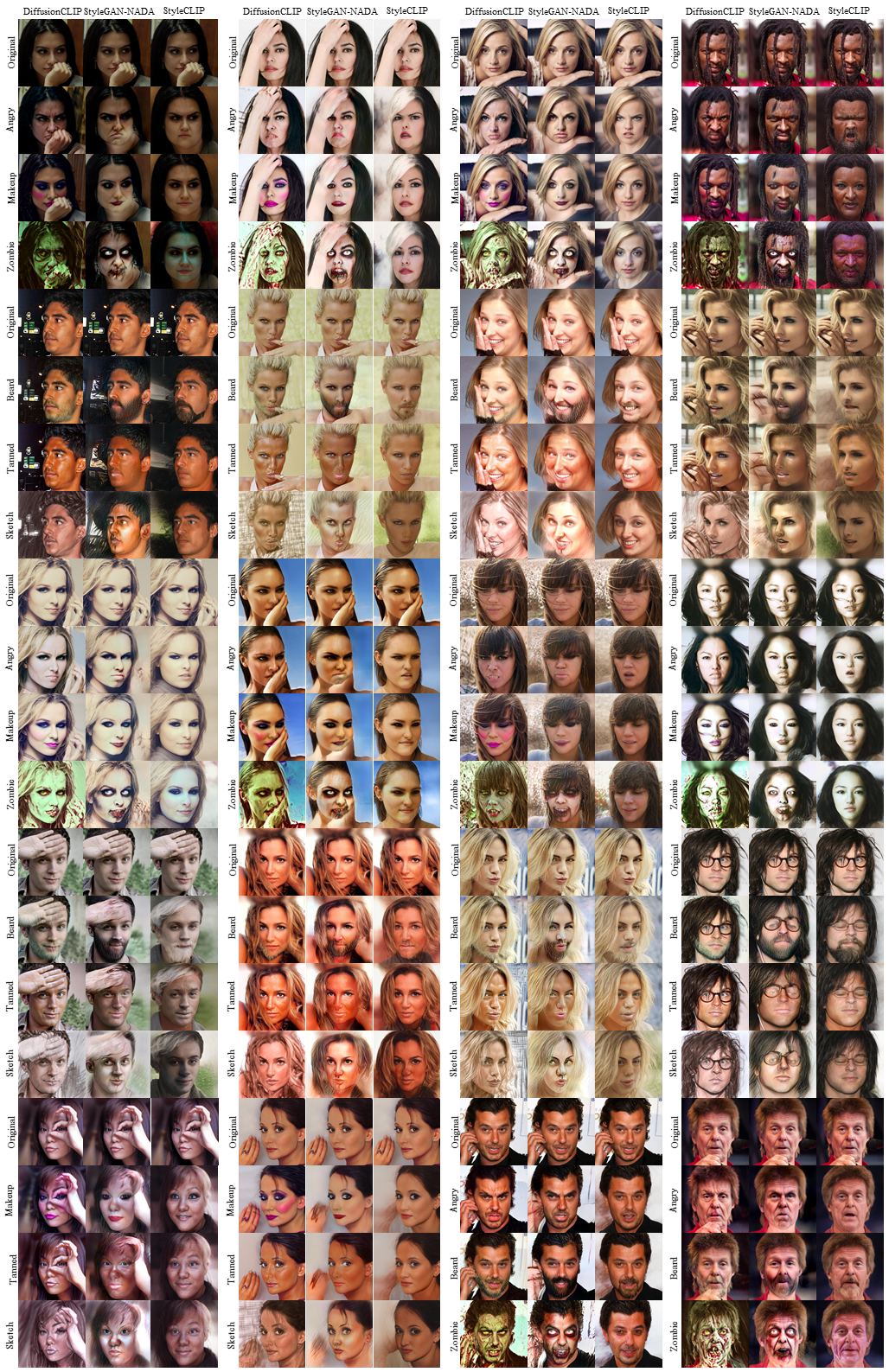}
    \vspace{-0.5em}
    \caption{Manipulation of hard cases that are used for human evaluation. Hard cases include 20 images with novel poses, views and details in CelebA-HQ~\cite{karras2017progressive}. We compare our method with StyleCLIP global direction method~\cite{patashnik2021styleclip} and StyleGAN-NADA~\cite{gal2021stylegan}.}
    \vspace{-1em}
    \label{fig_s:human_eval_hard}
\end{figure*}
\clearpage

\begin{figure*}[!htb]
    \centering
    \vspace{-2em}
    \includegraphics[width=0.85\linewidth]{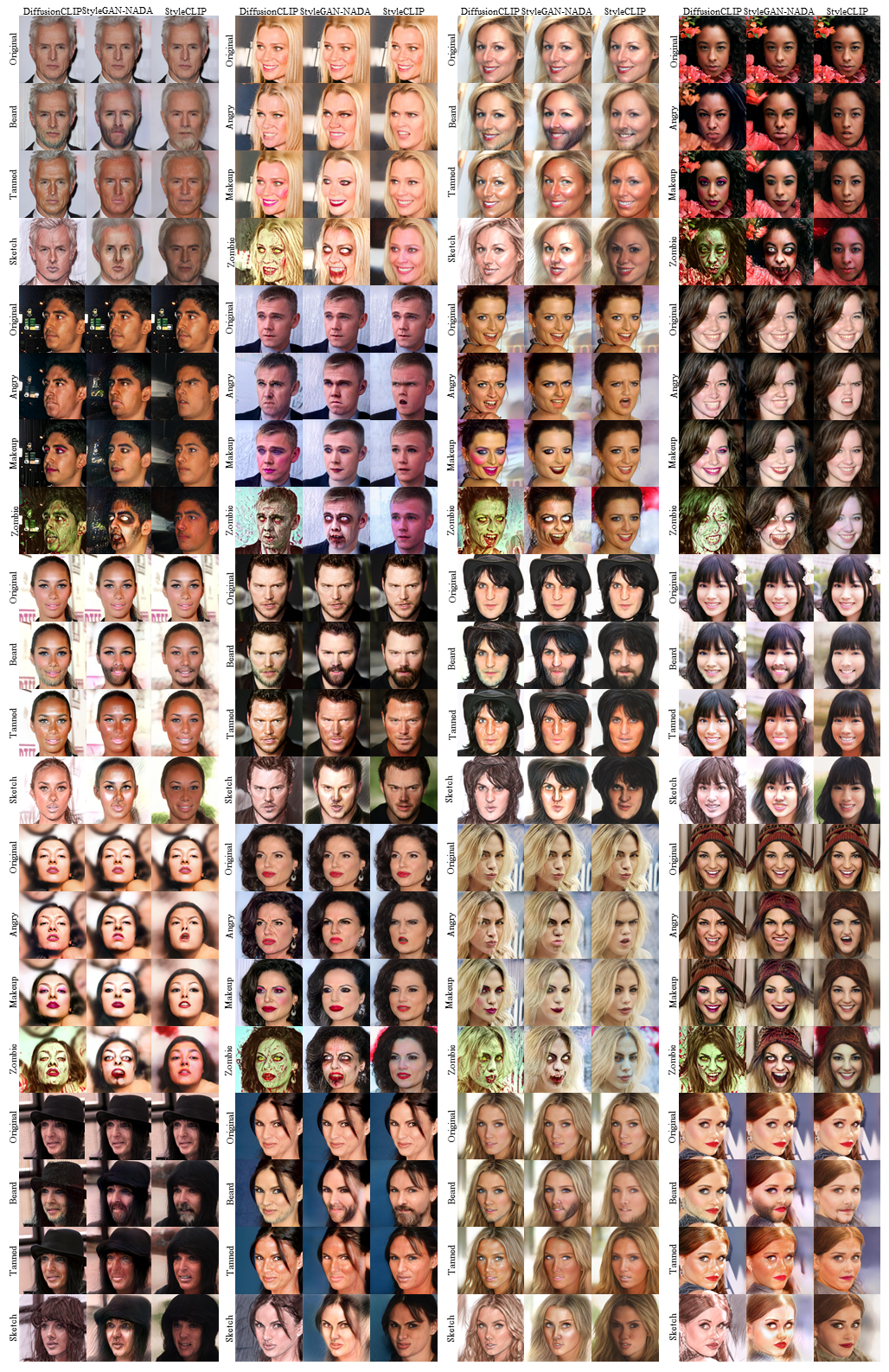}
    \vspace{-0.5em}
    \caption{Manipulation of general cases that are used for human evaluation. General cases include the first 20 images in CelebA-HQ testset~\cite{karras2017progressive}. We compare our method with StyleCLIP global direction method~\cite{patashnik2021styleclip} and StyleGAN-NADA~\cite{gal2021stylegan}.}
    \vspace{-1em}
    \label{fig_s:human_eval_general}
\end{figure*}
\clearpage

\begin{figure*}[!htb]
    \centering
    \vspace{-2em}
    \includegraphics[width=\linewidth]{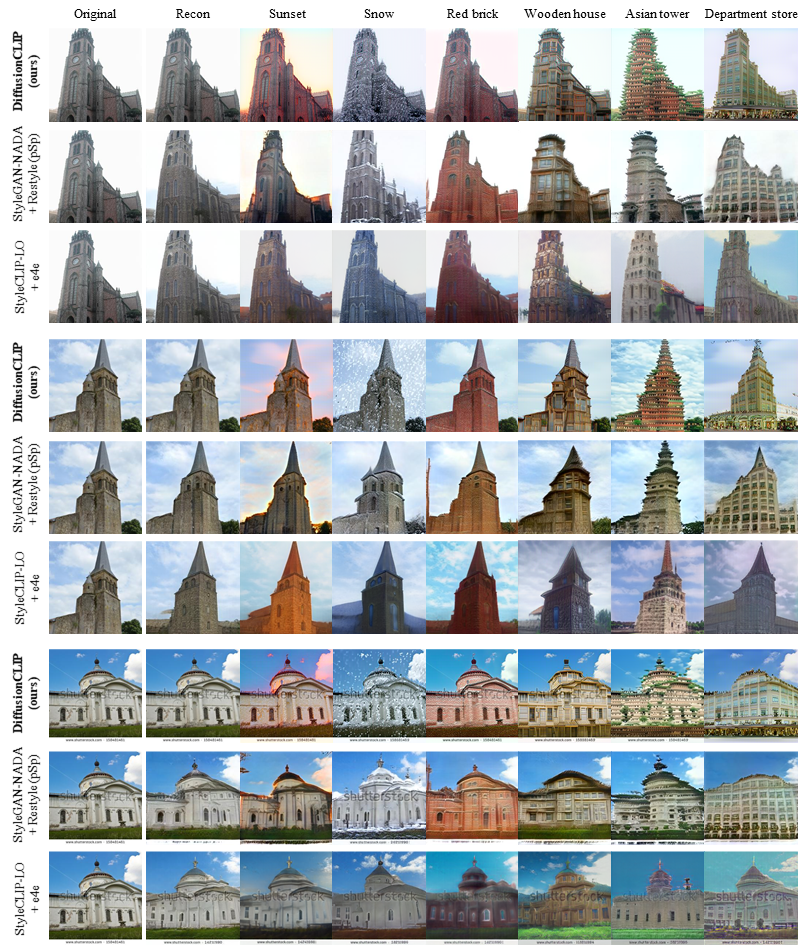}
    \caption{Qualitative comparison of church image manipulation performance with StyleCLIP global direction method~\cite{patashnik2021styleclip} and StyleGAN-NADA~\cite{gal2021stylegan}.}
    \vspace{-1em}
    \label{fig_s:comparision_church}
\end{figure*}
\clearpage

\clearpage
\begin{figure*}[!htb]
    \centering
    \vspace{-0.5em}
    \includegraphics[width=\linewidth]{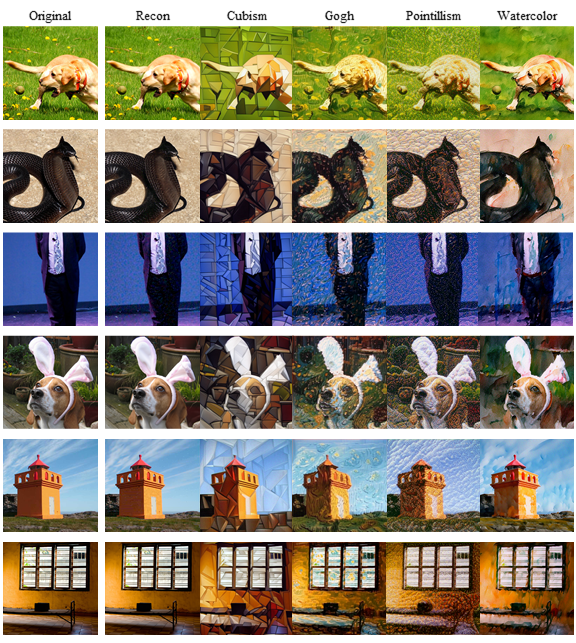}
    \caption{{Manipulation of $512\times512$ images using the ImageNet~\cite{ILSVRC15} pretrained diffusion models.}}
    \vspace{-1em}
    \label{fig_s:imagenet_general}
\end{figure*}

\clearpage
\begin{figure*}[!htb]
    \centering
    \vspace{-0.5em}
    \includegraphics[width=\linewidth]{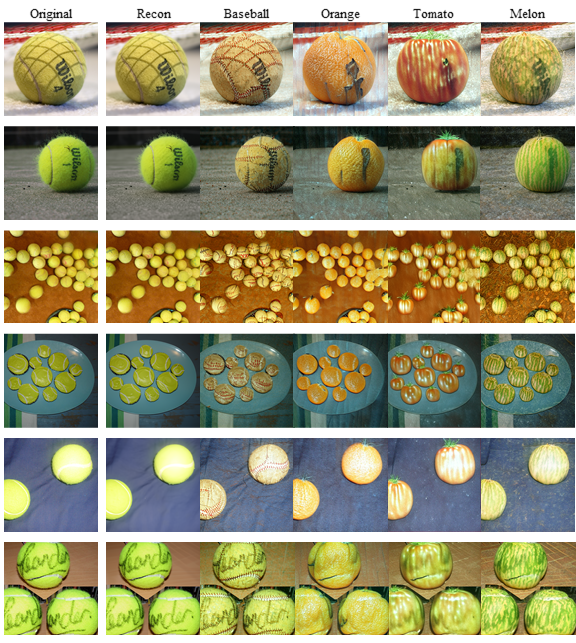}
    \caption{Manipulation of $512\times512$ images of tennis balls using the ImageNet~\cite{ILSVRC15} pretrained diffusion models.}
    \vspace{-1em}
    \label{fig_s:imagenet_tennis}
\end{figure*}

\clearpage
\begin{figure*}[!htb]
    \centering
    \vspace{-0.5em}
    \includegraphics[width=\linewidth]{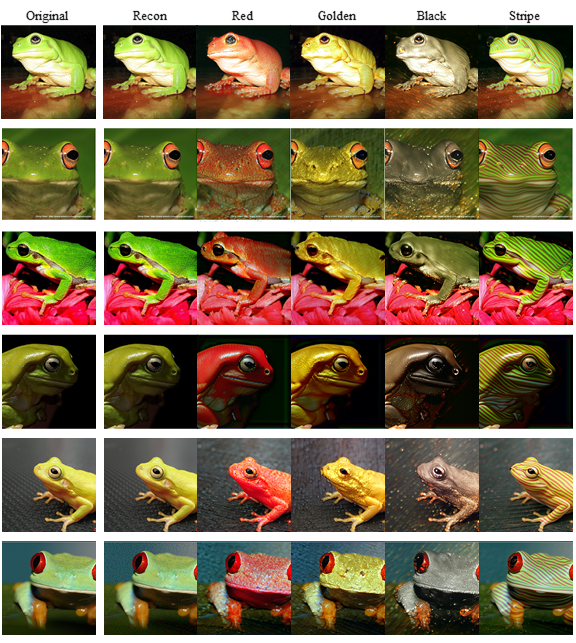}
    \caption{Manipulation of $512\times512$ images of frogs using the ImageNet~\cite{ILSVRC15} pretrained diffusion models.}
    \vspace{-1em}
    \label{fig_s:imagenet_frog}
\end{figure*}
\clearpage

\begin{figure*}[!htb]
    \centering
    \vspace{-2em}
    \includegraphics[width=\linewidth]{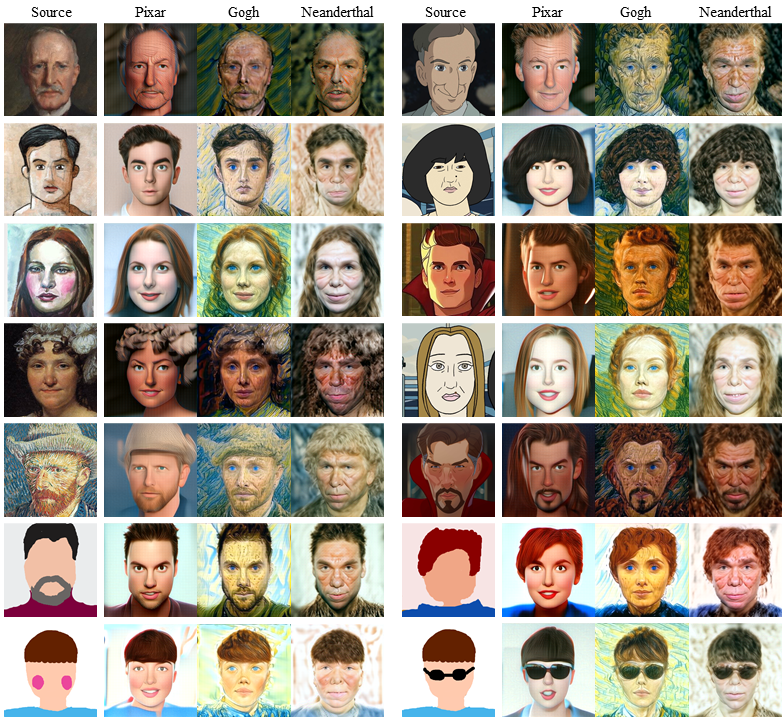}
    \vspace{-0.5em}
    \caption{Additional results of image translation between unseen domains. }
    \vspace{-1em}
    \label{fig_s:additional_unseen2unseen}
\end{figure*}
\clearpage

\end{document}

%% file: math_commands.tex
\usepackage{amsmath,amsfonts,bm}

\def\eqref#1{equation~\ref{#1}}

\def\1{\bm{1}}

\def\vf{{\bm{f}}}

\def\vx{{\bm{x}}}

\DeclareMathAlphabet{\mathsfit}{\encodingdefault}{\sfdefault}{m}{sl}
\SetMathAlphabet{\mathsfit}{bold}{\encodingdefault}{\sfdefault}{bx}{n}



%% file: macros.tex
\newcommand{\add}[1] {\textcolor{blue}{#1}} 

\newcommand{\x}{\boldsymbol{x}}

\newcommand{\y}{\boldsymbol{y}}

\newcommand{\fb}{{\boldsymbol f}}

\newcommand{\wb}{{\boldsymbol w}}
\newcommand{\xb}{{\boldsymbol x}}

\newcommand{\zb}{{\boldsymbol z}}

\newcommand{\Lc}{\mathcal{L}}

\newcommand{\mub}{{\boldsymbol{\mu}}}

\newcommand{\beq}{\begin{equation}}
\newcommand{\eeq}{\end{equation}}
\newcommand{\beqa}{\begin{eqnarray}}
\newcommand{\eeqa}{\end{eqnarray}}

\newcommand{\epsilonb}{\boldsymbol{\epsilon}}